%% file: main.tex
\documentclass[10pt,twocolumn,letterpaper]{article}
\usepackage[pagenumbers]{cvpr}

\definecolor{iccvblue}{rgb}{0.21,0.49,0.74}
\usepackage[pagebackref,breaklinks,colorlinks,allcolors=iccvblue]{hyperref}
\usepackage{multirow}

\title{Neighboring Autoregressive Modeling for Efficient Visual Generation}

\author{%
  Yefei He\textsuperscript{1,2}$^{\ast \dagger}$
  \quad  Yuanyu He\textsuperscript{1}$^{\ast}$
  \quad  Shaoxuan He\textsuperscript{1}$^{\ast}$
  \quad Feng Chen\textsuperscript{3}$^{\ast}$ \\
  \quad  Hong Zhou\textsuperscript{1}$^{\ddagger}$
  \quad  Kaipeng Zhang\textsuperscript{2}$^{\ddagger}$
  \quad  Bohan Zhuang\textsuperscript{1}$^{\ddagger}$
  \\[0.2cm]
  \textsuperscript{1}Zhejiang University, China \\
  \textsuperscript{2}Shanghai AI Laboratory, China \\
  \textsuperscript{3}The University of Adelaide, Australia \\
  \footnotesize{
  $^{\ast}$ Equal contribution \quad
  $\dagger$ Work done during an internship at Shanghai AI Laboratory \quad
  $^{\ddagger}$ Corresponding authors
  }
}

\begin{document}
\twocolumn[{
\renewcommand\twocolumn[1][]{#1}%
\maketitle
\vspace{-3em}
\begin{center}
    \centering
    \includegraphics[width=1.0\textwidth]{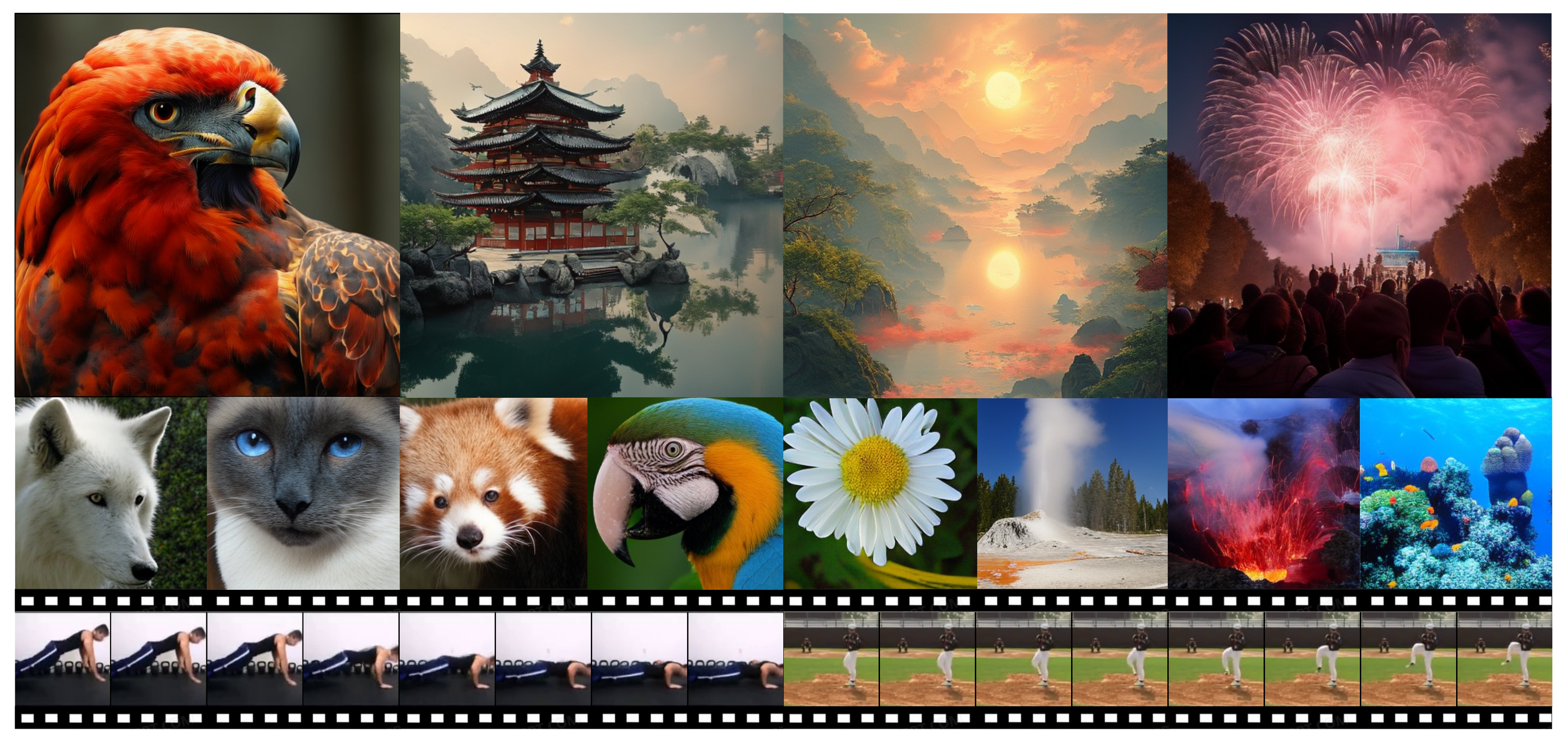}
    \captionof{figure}{\textbf{Generated samples from NAR.} Results are shown for $512\times512$ text-guided image generation (1st row), $256\times256$ class-conditional image generation (2nd row) and $128\times128$ class-conditional video generation (3rd row).}
    \label{fig:teaser}
\end{center}
}]

\input{sec/0_abstract}    
\input{sec/1_intro}
\input{sec/2_relatedworks}
\input{sec/3_method}
\input{sec/4_experiments}
\input{sec/5_conclusion}

{
    \small
    \bibliographystyle{ieeenat_fullname}
    \bibliography{main}
}
\newpage
\onecolumn
\input{sec/appendix}
\end{document}

%% file: sec/0_abstract.tex
\begin{abstract}
Visual autoregressive models typically adhere to a raster-order ``next-token prediction" paradigm, which overlooks the spatial and temporal locality inherent in visual content. 
Specifically, visual tokens exhibit significantly stronger correlations with their spatially or temporally adjacent tokens compared to those that are distant.
In this paper, we propose Neighboring Autoregressive Modeling (NAR), a novel paradigm that formulates autoregressive visual generation as a progressive outpainting procedure, following a near-to-far ``next-neighbor prediction" mechanism.
Starting from an initial token, the remaining tokens are decoded in ascending order of their Manhattan distance from the initial token in the spatial-temporal space, progressively expanding the boundary of the decoded region.
To enable parallel prediction of multiple adjacent tokens in the spatial-temporal space, we introduce a set of dimension-oriented decoding heads, each predicting the next token along a mutually orthogonal dimension.
During inference, all tokens adjacent to the decoded tokens are processed in parallel, substantially reducing the model forward steps for generation.
Experiments on ImageNet $256\times 256$ and UCF101 demonstrate that NAR achieves 2.4$\times$ and 8.6$\times$ higher throughput respectively, while obtaining superior FID/FVD scores for both image and video generation tasks compared to the PAR-4X approach.
When evaluating on text-to-image generation benchmark GenEval, NAR with 0.8B parameters outperforms Chameleon-7B while using merely \textbf{0.4\%} of the training data.
Code is available at \href{https://github.com/ThisisBillhe/NAR}{https://github.com/ThisisBillhe/NAR}.
\end{abstract}

%% file: sec/1_intro.tex
\section{Introduction}
Large language models (LLMs) \cite{radford2018improving,radford2019language,llama,llama2} trained with an autoregressive ``next-token prediction" objective have demonstrated unprecedented capabilities in addressing natural language-based tasks. Building on these advancements, many studies~\cite{llamagen,emu3,janus-pro, team2024chameleon, liu2024lumina} have explored autoregressive models for visual generation or even unified multimodal generation. 
Typically, to adapt the ``next-token prediction" paradigm for visual generation, images are tokenized into image tokens and flattened into one-dimensional visual token sequences in the raster order~\cite{esser2021taming,van2016pixelcnn,llamagen}. 
During inference, models must sequentially generate thousands of tokens to produce a single high-resolution image, resulting in significantly lower efficiency compared to diffusion-based models~\cite{sdxl,peebles2023scalable}.

\begin{figure*}[t]
\centering
\includegraphics[width=0.9\textwidth]{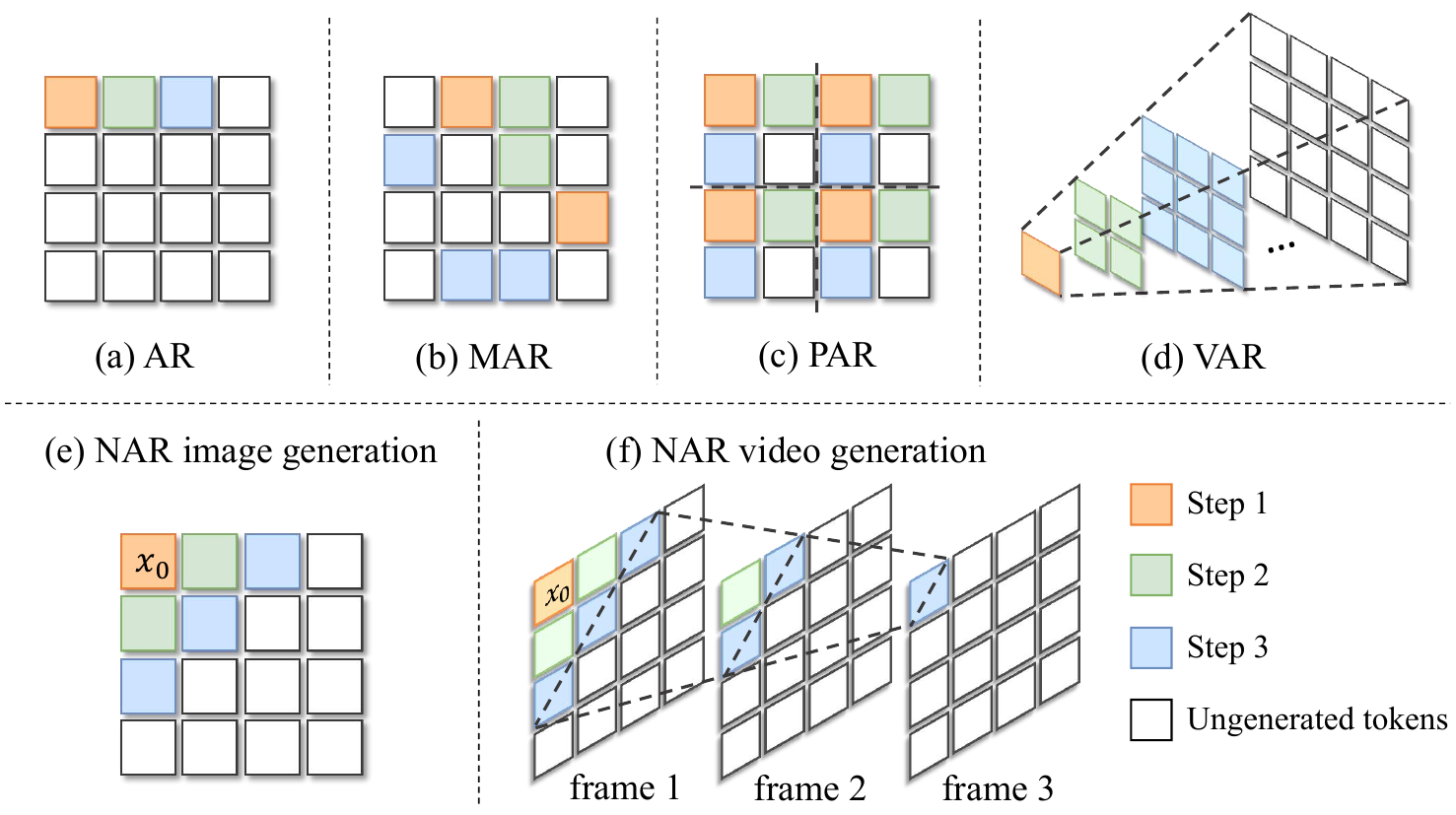}
\caption{ 
\textbf{Comparisons of different autoregressive visual generation paradigm.} The proposed NAR paradigm formulates the generation process as an outpainting procedure, progressively expanding the boundary of the decoded token region. This approach effectively preserves locality, as all tokens near the starting point are consistently decoded before the current token.}
\label{fig:paradigm_comparison}
\end{figure*}

Recent efforts to accelerate autoregressive visual generation can be broadly classified into two categories. 
The first category parallelly generates multiple tokens in an image within one step. For instance, SJD~\cite{jacobi} retains a raster-order visual sequence and predicts multiple consecutive tokens in each forward step. 
However, prior studies~\cite{he2024zipar, tian2024var} have demonstrated that image tokens exhibit strong correlations with their spatially adjacent counterparts. Consequently, parallel decoding of consecutive tokens leads to insufficient context and notable performance degradation, particularly as the level of parallelism increases.
Similarly, MAR~\cite{li2024mar} and PAR~\cite{wang2024parallelized} generate multiple tokens in randomly selected positions or block-constrained regions, as shown in Figure~\ref{fig:paradigm_comparison}(b)-(c). These methods require extensive hyperparameter tuning to balance parallelism and spatial coherence.
The second category follows a ``next-scale prediction" paradigm~\cite{tian2024var,li2024controlvar}, where token maps are autoregressively generated from coarse to fine scales, as illustrated in Figure~\ref{fig:paradigm_comparison}(d).
However, their token sequence is constructed by concatenating visual tokens from multiple scales. This requires specialized multi-scale image tokenizers and increases the overall length compared to single-scale visual sequences. As a result, it leads to higher memory overhead during both training and inference.
To establish an efficient visual autoregressive modeling paradigm, we argue that it should satisfy the following criteria: 1) preserve spatial or temporal locality within visual features; 2) support parallel decoding during inference; 3) be compatible with regular single-scale tokenizers and maintain a short visual token sequence.
Unfortunately, none have yet achieved all of these objectives simultaneously.

\begin{figure}[t]
\includegraphics[width=0.5\textwidth]{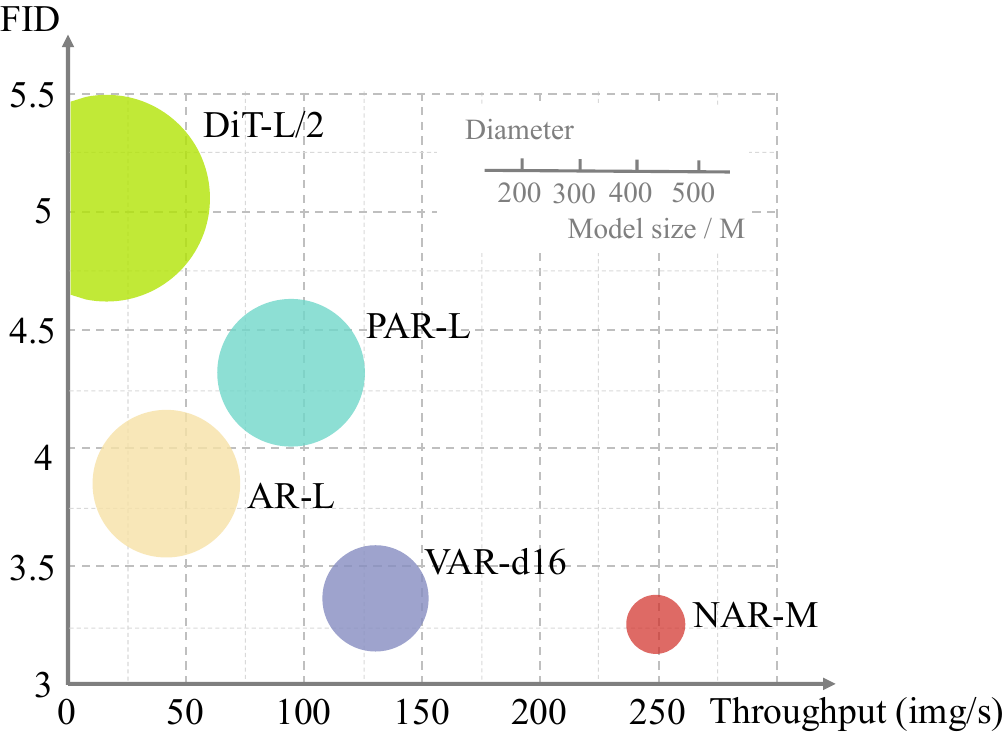}
\caption{ 
Generation quality and efficiency comparisons between various visual generation methods. Data is collected from ImageNet $256\times256$ dataset over models with parameters around 300M.
}
\label{fig:efficiency_overview}
\end{figure}

In this paper, we propose neighboring autoregressive modeling (NAR), a novel paradigm for high-fidelity and efficient visual generation.  Figure~\ref{fig:efficiency_overview} presents an overview of generation
efficiency and quality comparisons among NAR and diverse visual generation methods. 
Drawing inspiration from image outpainting approaches~\cite{lin2021edge, sabini2018painting}, which extend the content of an image beyond its original boundaries, we frame autoregressive visual generation as a progressive outpainting process that begins from an initial token and is guided by a near-to-far 'next-neighbor prediction' mechanism.
As illustrated in Figure~\ref{fig:paradigm_comparison}(e), our approach begins by arranging visual tokens in a locality-preserved order. Specifically, the Manhattan distance between visual tokens and the top left token in the spatial domain corresponds to their generation order. This arrangement poses a significant challenge for models based on the ``next-token prediction" paradigm, as multiple equidistant tokens can exist in different dimensions, yet only one token can be generated at each step.
To address this, we introduce dimension-oriented decoding heads, each responsible for predicting the next token along a distinct, mutually orthogonal dimension in the spatial space. During inference, multiple dimension-oriented decoding heads naturally support parallel decoding. Once a set of visual tokens has been decoded, all adjacent tokens can be generated in the subsequent step, thereby enabling the ``next-neighbor prediction" paradigm and significantly improving generation efficiency. 
This approach can be seamlessly extended to video generation by employing three decoding heads and applying the same strategy, as demonstrated in Figure~\ref{fig:paradigm_comparison}(f).
Extensive experiments on image and video generation demonstrate that visual generation with the next-neighbor prediction paradigm can greatly improve the efficiency of visual generation while producing superior results compared to models with the next-token prediction paradigm. For class-conditional image generation on ImageNet $256 \times 256$, our method \textbf{reduces the number of generation steps by $91.8\%$ while lowering the FID by $0.81$} over LlamaGen-XL model. For class-conditional video generation on UCF-101, our method achieves a competitive \textbf{FVD of $71.1$ with a $97.3\%$ generation step reduction} compared to the vanilla next-token AR methods. 
When evaluating text-guided image generation on the GenEval benchmark, NAR-0.8B with merely 6M training data outperforms Chameleon-7B with 1.4B training data, demonstrating the effectiveness of our approach in generating high-resolution, high-aesthetic images.

In summary, our contributions are as follows:

\begin{itemize}
\item We propose neighboring autoregressive modeling (NAR), a new next-neighbor prediction paradigm that formulates autoregressive visual generation as a near-to-far outpainting process.

\item We introduce dimension-oriented decoding heads, each responsible for predicting the next token along distinct, mutually orthogonal dimensions in the spatial-temporal space.
This innovation allows for the parallel generation of all adjacent tokens in one step, enhancing the generation efficiency for both images and videos.

\item  Extensive experiments on class-conditional image generation, class-conditional video generation, and text-guided image generation demonstrate that our method achieves substantial efficiency gains and superior generation quality compared to the existing autoregressive visual generation methods.
\end{itemize}

%% file: sec/2_relatedworks.tex
\section{Related Work}
\subsection{Efficient Autoregressive Visual Generation} 
Typically, autoregressive (AR) visual generation models trained with a next-token prediction objective, necessitate $n^2$ sequential model forward passes to generate an image represented by $n\times n$ tokens, resulting in significantly low generation efficiency. This issue is even more pronounced for high-resolution image~\cite{liu2024lumina} or video generation~\cite{emu3,deng2024nova}, hindering its wide deployment in real-world applications. 

To address this inefficiency, techniques such as SJD~\cite{jacobi} and ZipAR~\cite{he2024zipar} have been proposed as training-free methods to accelerate sampling by predicting multiple consecutive or spatially adjacent tokens in a single step. However, these methods provide only limited speedup, as the pretrained AR models are designed solely to model the distribution of the next token in raster order.
Another approach, PAR~\cite{wang2024parallelized}, divides image tokens into spatially distant subsets and employs the next-token prediction paradigm within each subset for parallel generation. Although PAR trains models from scratch, its performance may be suboptimal due to insufficient global context when generating multiple distant regions simultaneously.
In contrast to the traditional next-token prediction paradigm, MAR~\cite{li2024mar} predicts multiple output tokens simultaneously in a random order. VAR~\cite{tian2024var} iteratively generates a coarse-to-fine multi-resolution image token map using a next-scale prediction paradigm. However, VAR relies on multi-scale image tokenizers and produces longer token sequences, complicating the process and increasing training overhead.
In this paper, we propose an efficient neighboring autoregressive generation paradigm that retains the short token sequences and image tokenizers of vanilla next-token AR models while improving image quality and achieving significantly higher generation efficiency.

\subsection{Visual Locality in Computer Vision}
The concept of visual locality refers to the similarity among adjacent pixels in terms of color, texture, and edges. This locality has been extensively investigated in early computer vision research, such as image compression~\cite{wallace1991jpeg, boutell1997png}, image denoising~\cite{tomasi1998bilateral, dabov2007imagedenoising}, and edge detection~\cite{Sobel1973,canny1986computational}.
Convolutional neural networks (CNNs)~\cite{lenet,krizhevsky2012imagenet,simonyan2014very,he2015deep} leverage this property by restricting interactions to local neighborhoods, which has enabled them to significantly outperform multi-layer perceptron (MLP) models across a wide range of computer vision tasks~\cite{redmon2015you,long2014fully,he2017mask}.
In contrast, the standard transformer model~\cite{waswani2017attention} utilizes a self-attention mechanism to model interactions between tokens, which inherently lacks a focus on local patterns. This limitation leads to suboptimal performance when such models are directly applied to image data~\cite{dosovitskiy2020vit}. To address this issue, several techniques have been proposed, including the integration of CNNs with self-attention models~\cite{bello2020attnaug,carion2020endtoend,xi2017pixelsnail} and the implementation of self-attention within localized windows~\cite{swin, zhang2022nested}. These methods aim to simultaneously capture fine-grained visual locality and long-range dependencies.
Despite these advancements, the application of spatial locality in autoregressive image generation remains unexplored. In this paper, we introduce a locality-preserved generation paradigm, which demonstrates superior efficiency and quality in visual generation tasks.

%% file: sec/3_method.tex
\begin{figure}
\centering
\includegraphics[width=0.5\textwidth]{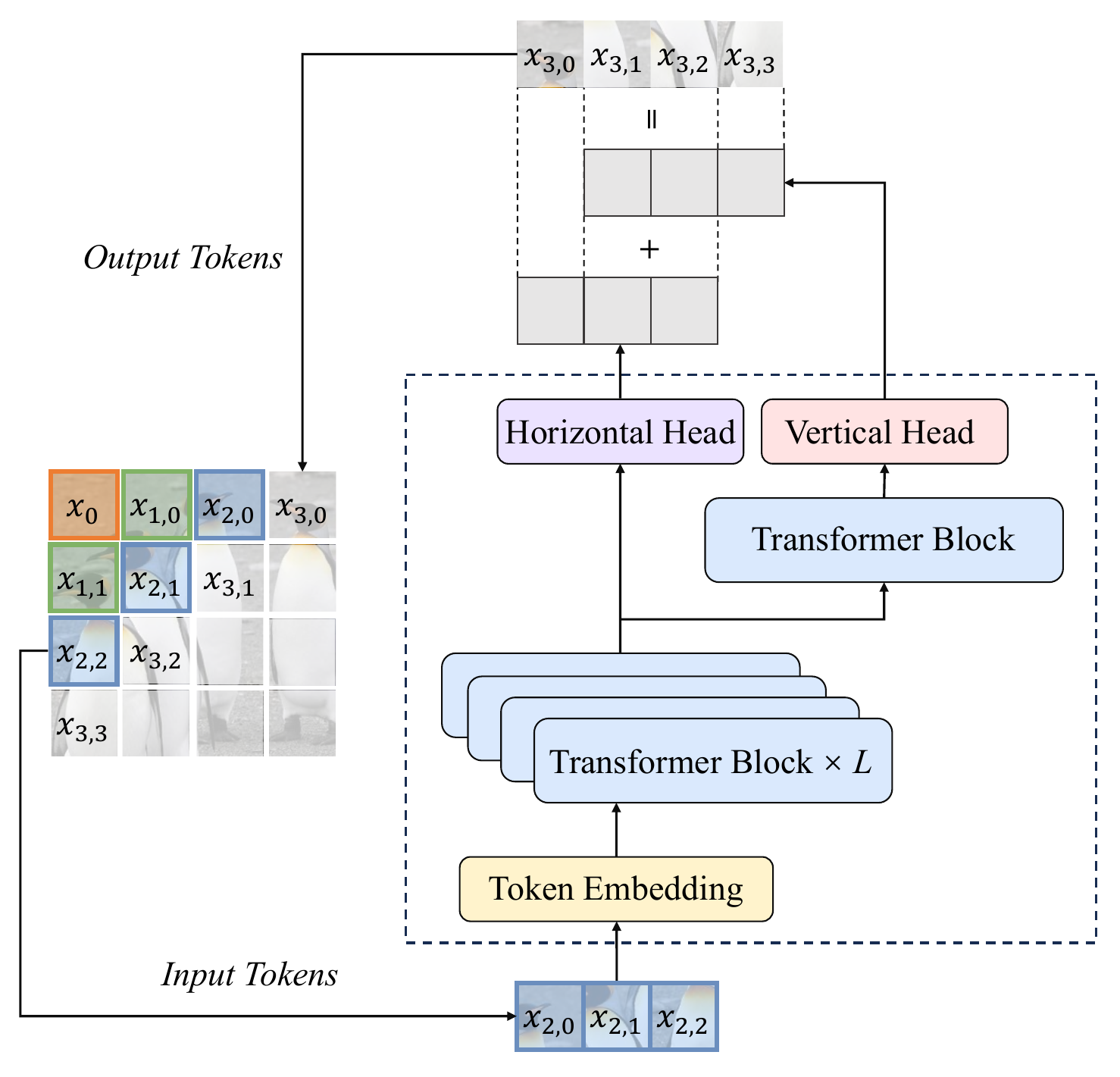}
\caption{
\textbf{Illustration of the dimension-oriented decoding heads.} The horizontal head and the vertical head are responsible for predicting the next token in the row and column dimensions, respectively. Here, $L$ is the number of Transformer blocks in the backbone network.
}
\label{fig:architecture}
\end{figure}

\section{Method}
\subsection{Neighboring Autoregressive Modeling}
Motivated by image outpainting methods~\cite{lin2021edge, sabini2018painting}, we propose a novel perspective on autoregressive visual generation. As shown in Figure~\ref{fig:paradigm_comparison}(e)-(f), our approach frames the process as an outpainting procedure initiated from scratch, progressively expanding the boundary of the decoded token region. At each generation step, all adjacent tokens to the decoded tokens are predicted, leading to a neighboring autoregressive modeling (NAR) paradigm. This paradigm effectively preserves locality, as all tokens near the starting point are consistently decoded before the current token.

However, this configuration presents a significant challenge for models trained with the next-token prediction (NTP) objective, as multiple equidistant tokens may exist in different dimensions, yet only one token can be generated per step. Although certain methods~\cite{jacobi, he2024zipar, wang2024parallelized} enable parallel decoding within the NTP framework, their performance significantly degrades when predicting too many tokens simultaneously. 
As evidenced by the empirical results in Table~\ref{tab:ablate_heads}, directly integrating the NAR paradigm with the next-token prediction objective leads to substantial performance degradation.
We contend that this issue arises because the original model is trained exclusively to predict the conditional distribution of the next token in raster order, while the distribution of tokens across multiple locations to be predicted can vary substantially.

To address this, we propose dimension-oriented decoding heads, a simple yet effective approach to model the distinct conditional distributions of parallel decoded tokens. As presented in Figure~\ref{fig:architecture}, the extra decoding heads consist of a Transformer block and a fully-connected output layer, which are concatenated to the Transformer backbone to enable parallel decoding along several mutually orthogonal dimensions. For instance, an image can be considered two-dimensional, allowing decoding to proceed along two orthogonal dimensions: rows and columns. Consequently, we apply two dimension-oriented decoding heads for image generation models, with each head modeling the conditional distribution of the next token in one dimension, i.e., the next token in the same row and the token in the same column in the next row.
Our method can also be naturally extended to video generation. Since videos can be regarded as three-dimensional, adding a temporal dimension to images, decoding can be performed along three orthogonal dimensions: times, rows, and columns. Such a decoupled design prevents a single head from predicting multiple mixed conditional distributions, thereby significantly improving the generation performance.

\begin{figure}
\centering
\includegraphics[width=0.5\textwidth]{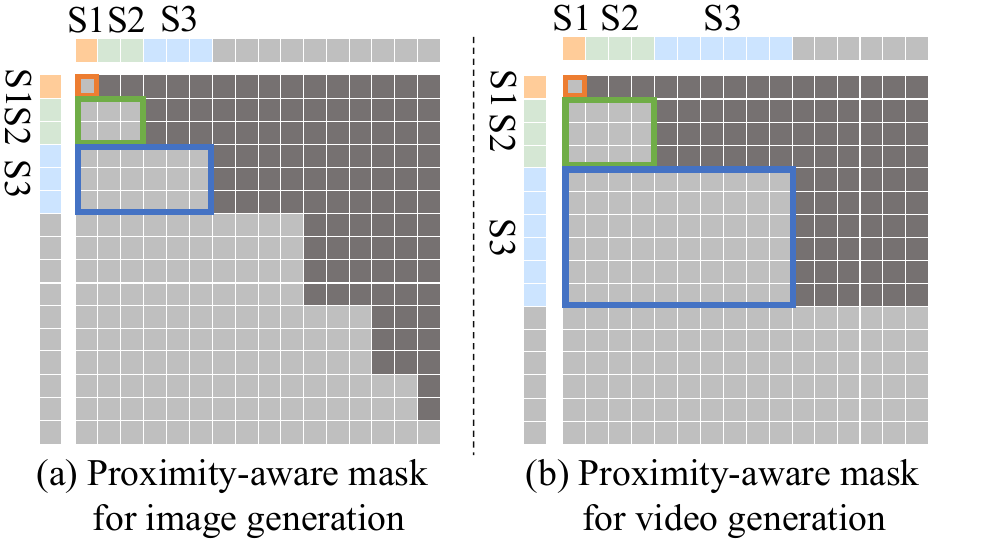}
\caption{\textbf{Proximity-aware attention masks for the NAR paradigm.} ``S$n$'' denotes the $n$-th generation step. Tokens generated within the same step are represented by the same color. To maintain the autoregressive property, a causal mask is applied between tokens across different generation steps (aligned with Figure~\ref{fig:paradigm_comparison}). Within each step, bidirectional attention is employed among the tokens to enhance consistency during parallel generation.}
\label{fig:mask}
\end{figure}

\subsection{Implementation Details of NAR}

\noindent \textbf{Inference with NAR.}
By employing a set of decoding heads to predict the next token along multiple dimensions, we can naturally predict all adjacent tokens of a generated token in one forward step. Specifically, the decoding process begins with the initial token positioned at the upper-left corner of the image feature map, aligning with the raster-scan order. In subsequent steps, the tokens generated in the previous step are input into the model to autoregressively generate new tokens, as illustrated in Figure~\ref{fig:architecture}. For each input token, its two adjacent tokens along the row and column dimensions are generated using dimension-oriented heads. Notably, starting from the third step, there are overlaps between the two predicted neighboring tokens for input tokens, \eg, $x_{3,1}$ and $x_{3,2}$ in Figure~\ref{fig:architecture}. In this case, we mix the predictions from different decoding heads for the overlapped tokens, analogous to a model ensemble approach~\cite{lu2024ensemble}. Empirical results, presented in Table~\ref{tab:mixlogits}, demonstrate that this method consistently enhances generation performance compared to relying on the prediction of a single decoding head.

Formally, given the initial token $x_0$, the set of all tokens generated at step $i$ can be defined as:
\begin{align}
    S = \{ x_i \mid D(x_i, x_0) = i \},
\end{align}
where $D$ represents the Manhattan distance. 
With such a generation paradigm, NAR produces a high-resolution image with \( n \times n \) tokens in $2n-1$ steps. For videos represented by  \( t \times n \times n \) tokens, \textbf{NAR completes the generation process in only $2n+t-2$ steps}, significantly fewer than the $tn^2$ steps required by vanilla next-token AR models.

\noindent \textbf{Training with NAR.}
Training a NAR model shares the same image tokenizer and training pipeline as vanilla next-token AR models, requiring only minor modifications to the model architecture and the attention mask. 
% Firstly, a set of dimension-oriented decoding heads are concatenated after the Transformer backbone. 
Similar to previous AR image generation methods~\cite{llamagen,tian2024var}, all decoding heads are trained with cross-entropy loss, predicting the next token along different dimensions. 
Since the decoding order of tokens is determined by their proximity to the initial token, tokens farther from the initial token depend only on those closer to it during decoding. Therefore, a proximity-aware causal attention mask is employed for training NAR models, as shown in Figure~\ref{fig:mask}. Notably, bidirectional attention is applied among tokens equidistant from the initial token, ensuring better consistency during parallel generation.

Compared to the VAR paradigm, the proposed NAR paradigm substantially reduces training overhead. First, VAR necessitates a specialized multi-scale tokenizer, which requires extensive data for training to achieve optimal performance. In contrast, our approach can leverage a diverse range of high performance image tokenizers from the open-source community. Second, the token sequence of VAR is constructed by concatenating visual tokens from multiple scales, which is considerably longer than the single-scale visual token sequence used in NAR. This reduction in sequence length substantially decreases computational complexity during training, as the complexity scales quadratically with the sequence length.

%% file: sec/4_experiments.tex
\begin{table*}[h] \small
\caption{Quantitative evaluation on the ImageNet $256\times256$ benchmark. ``Step'' denotes the number of model forward passes required to generate an image. The throughput is measured with the maximum batch size supported on a single A100 GPU. Classifier-free guidance is set to 2 for our method. We also report the reconstruction FID (rFID) of visual tokenizers for each method, which serves as an upper bound for generation FID. $\dagger$: model denoted as M shares the same hidden dimension as the L model but is reduced by 6 layers in depth.}
\label{tab:imagenet}
\centering
\scalebox{1.0}{
\begin{tabular}{cccccccccc}
\hline
Tokenizer                                                                         & Type                 & Model        & Params & FID$\downarrow$ & IS$\uparrow$   & Precision$\uparrow$ & Recall$\uparrow$ & Steps & \begin{tabular}[c]{@{}c@{}}Throughput\\ (img/s)\end{tabular} \\ \hline
\multirow{4}{*}{\begin{tabular}[c]{@{}c@{}}VQVAE-108M\\ (rFID=1.00)\end{tabular}} & \multirow{4}{*}{VAR} & VAR-d16      & 310M   & 3.30            & 274.4          & \textbf{0.84}       & 0.51             & 10    & \textbf{129.3}                                               \\
                                                                                  &                      & VAR-d20      & 600M   & 2.57            & 302.6          & 0.83                & 0.56             & 10    & 77.6                                                         \\
                                                                                  &                      & VAR-d24      & 1.0B   & 2.09            & 312.9          & 0.82                & 0.59             & 10    & 50.6                                                         \\
                                                                                  &                      & VAR-d30      & 2.0B   & \textbf{1.92}   & \textbf{323.1} & 0.82                & \textbf{0.59}    & 10    & 29.9                                                         \\ \hline
\multirow{13}{*}{\begin{tabular}[c]{@{}c@{}}VQVAE-72M\\ (rFID=2.19)\end{tabular}} & \multirow{4}{*}{AR}  & LlamaGen-B   & 111M   & 5.46            & 193.6          & 0.84                & 0.46             & 256   & 117.9                                                        \\
                                                                                  &                      & LlamaGen-L   & 343M   & 3.80            & 248.3          & 0.83                & 0.52             & 256   & 47.1                                                         \\
                                                                                  &                      & LlamaGen-XL  & 775M   & 3.39            & 227.1          & 0.81                & 0.54             & 256   & 23.7                                                         \\
                                                                                  &                      & LlamaGen-XXL & 1.4B   & 3.09            & 253.6          & 0.83                & 0.53             & 256   & 14.1                                                         \\ \cline{2-10} 
                                                                                  & \multirow{4}{*}{PAR} & PAR-B-4X     & 111M   & 6.21            & 204.4          & 0.86                & 0.39             & 67    & 174.1                                                        \\
                                                                                  &                      & PAR-L-4X     & 343M   & 4.32            & 189.4          & \textbf{0.87}       & 0.43             & 67    & 93.8                                                         \\
                                                                                  &                      & PAR-XL-4X    & 775M   & 3.50            & 234.4          & 0.84                & 0.49             & 67    & 53.9                                                         \\
                                                                                  &                      & PAR-XXL-4X   & 1.4B   & 3.20            & 288.3          & 0.86                & 0.50             & 67    & 33.9                                                         \\ \cline{2-10} 
                                                                                  & \multirow{5}{*}{NAR} & NAR-B        & 130M   & 4.65            & 212.3          & 0.83                & 0.47             & 31    & \textbf{419.7}                                               \\
                                                                                  &                      & NAR-M$\dagger$        & 290M   & 3.27            & 257.5          & 0.82                & 0.53             & 31    & 248.5                                                        \\
                                                                                  &                      & NAR-L        & 372M   & 3.06            & 263.9          & 0.81                & 0.53             & 31    & 195.4                                                        \\
                                                                                  &                      & NAR-XL       & 816M   & 2.70            & 277.5          & 0.81                & \textbf{0.58}    & 31    & 98.1                                                         \\
                                                                                  &                      & NAR-XXL      & 1.46B  & \textbf{2.58}   & \textbf{293.5} & 0.82                & 0.57             & 31    & 56.9                                                         \\ \hline
\end{tabular}}
\end{table*}

\section{Experiments}
\subsection{Experimental Setup}
\noindent \textbf{Model architecture.} To validate the effectiveness and scalability of the proposed NAR paradigm, we adopt a decoder-only Transformers architecture, following prior studies~\cite{llamagen, wang2024parallelized,wang2024larp}. 
By implementing dimension-oriented decoding heads, our NAR is inherently applicable to both two-dimensional image and three-dimensional video generation,
utilizing two and three decoding heads, respectively.

\noindent \textbf{Class-conditional image generation.} We evaluate NAR on the widely adopted ImageNet $256\times256$ dataset. Images are tokenized using an off-the-shelf image tokenizer introduced by~\cite{llamagen}, with a downsampling factor of $16$. All models are trained with 300 epochs with a base learning rate of $10^{-4}$ and a step learning rate scheduler. The reported Inception Score (IS) and Frechet Inception Distance (FID) results are computed by sampling 50,000 images and evaluating them with ADM’s TensorFlow evaluation suite~\cite{dhariwal2021adm}. 

\noindent \textbf{Class-conditional video generation.} NAR is trained and evaluated on UCF-101 dataset~\cite{soomro2012ucf101}. We utilize a video tokenizer proposed by~\cite{wang2024larp}, which encodes a $16\times128\times128$ video clip into $4\times16\times16$ visual tokens. Models are trained with 3000 epochs with a base learning rate of $10^{-4}$ and step learning rate scheduler. Frechet Video Distance (FVD)~\cite{unterthiner2018towards} serves as the main evaluation metric for generation.

\noindent \textbf{Text to image generation.} We curate a dataset comprising 4M image-text pairs from the LAION-COCO~\cite{laioncoco2023} dataset and 2M open-sourced high-resolution images captioned using large vision-language models~\cite{wang2024qwen2vl}. An image tokenizer fine-tuned on LAION-COCO, introduced by~\cite{llamagen}, is employed with a downsampling factor of $16$. Text embeddings are extracted using the pretrained FLAN-T5 model~\cite{chung2024flant5}, which serves as the conditional input for image generation. Following prior work~\cite{llamagen}, the training process is divided into two stages. In the first stage, the model is trained on the 4M LAION-COCO subset at a resolution of $256\times256$ for 60 epochs. In the second stage, the model is fine-tuned on the 2M high-quality dataset at a resolution of $512\times512$ for 40 epochs. A cosine-annealing learning rate scheduler is used for both training stages. GenEval~\cite{ghosh2024geneval} is adopted as a fair and fine-grained benchmark for comparison.

\subsection{Main Results}
\subsubsection{Class-conditional Image Generation}

In this subsection, we evaluate the performance of NAR models on ImageNet $256\times256$ dataset, as presented in Table~\ref{tab:imagenet}. For a fair comparison, we adopt the same image tokenizer, model architectures, and training pipelines as LlamaGen~\cite{llamagen} and PAR~\cite{wang2024parallelized}. The employed image tokenizer has merely 72M parameters, which is exclusively trained on the ImageNet dataset and provides a reconstruction FID (rFID) of 2.19.
Although the previous parallel decoding method, PAR, improves generation efficiency compared to LlamaGen's standard next-token prediction paradigm, its FID remains consistently higher than that of LlamaGen at the same model size. 
In contrast, models employing the NAR paradigm exhibit superior performance and efficiency.
For instance, \textbf{NAR-L, with 372M parameters, achieves a lower FID than LlamaGen-XXL with 1.4B parameters (3.06 \emph{vs.} 3.09)}, while reducing the number of model forward steps by 87.8\% (31 steps \emph{vs.} 256 steps) and \textbf{delivering 13.8$\times$ higher throughput (195.4 images/s \emph{vs.} 14.1 images/s).} 

On the other hand, the VAR approach~\cite{tian2024var} employs a larger image tokenizer with 108M parameters, which is trained on the large-scale OpenImages dataset~\cite{kuznetsova2020openimages}. This tokenizer provides a reconstruction FID (rFID) of 1.00 and offers a higher upper bound for generation performance. Despite this, NAR-M with fewer parameters achieves \textbf{lower FID than VAR-d16 (3.27 \emph{vs.} 3.30)}, while providing \textbf{1.92$\times$ higher throughput (248.5 images/s \emph{vs.} 129.3 images/s).} Combining NAR with a more advanced image tokenizer will be left as future work.

\subsubsection{Class-conditional Video Generation}

In this subsection, we evaluate the effectiveness of NAR on class-based video generation using UCF-101~\cite{soomro2012ucf101}. As shown in Table~\ref{tab:ucf101}, our NAR surpasses other autoregressive counterparts by significantly reducing generation steps and wall-clock time while achieving lower FVD. Compared to LARP-L-Long~\cite{wang2024larp}, which employs the same video tokenizer and has a comparable number of parameters, our NAR-L further enhances generation quality, \textbf{reducing FVD by 5.8 and generation latency by 97.5\%.} Furthermore, compared to PAR~\cite{wang2024parallelized}, which is designed for parallel generation, our NAR-XL consistently outperforms it with a 23$\sim$32.3 FVD reduction, all without requiring hyperparameter tuning. Overall, we enhance the scalability of the autoregressive paradigm for image/video generation, making it comparable to diffusion-~\cite{luo2023videofusion, singer2022make} and masking-based~\cite{yu2023magvit,yu2023language} approaches while using fewer parameters and incurring lower latency. 

\begin{table}[t]
\caption{Comparison of class-conditional video generation methods on UCF-101 benchmark. Classifier-free guidance is set to 1.25 for all variants of our method. $\dagger$: model denoted as LP shares the same hidden dimension as the XL model but is reduced by 6 layers in depth.}
\label{tab:ucf101}
\centering
\scriptsize{
\begin{tabular}{c|l|c|c|c|c}
\toprule
Type & Method & Params & FVD$\downarrow$ & Steps & Time (s) \\ 
\midrule
\multirow{3}{*}{Diffusion} 
& VideoFusion~\cite{luo2023videofusion}      & N/A   & 173   & -    & -    \\ 
& Make-A-Video~\cite{singer2022make}     & N/A   & 81.3  & -    & -    \\ 
& HPDM-L~\cite{skorokhodov2024hierarchical}           & 725M  & 66.3  & -    & -    \\ 
\midrule
\multirow{2}{*}{Masking} 
& MAGVIT~\cite{yu2023magvit}           & 306M  & 76    & -    & -    \\ 
& MAGVIT-v2~\cite{yu2023language}        & 840M  & \textbf{58}    & -    & -    \\ 
\midrule
\multirow{8}{*}{AR} 
& CogVideo~\cite{hong2022cogvideo}         & 9.4B  & 626   & -    & -    \\ 
& TATS~\cite{ge2022long}             & 321M  & 332   & -    & -    \\ 
& OmniTokenizer~\cite{wang2024omnitokenizer}    & 650M  & 191   & 5120 & 336.70 \\ 
& MAGVIT-v2-AR~\cite{yu2023language}     & 840M  & 109   & 1280 & -     \\ 
& LARP-L-Long~\cite{wang2024larp}     &  343M  & 102   &  1280 & 44.0     \\
\cmidrule(l){2-6}
& PAR-XL-1×~\cite{wang2024parallelized}          & 792M  & 94.1  & 1280 & 43.30 \\ 
& PAR-XL-4×~\cite{wang2024parallelized}            & 792M  & 99.5  & 323  & 11.27 \\ 
& PAR-XL-16×~\cite{wang2024parallelized}           & 792M  & 103.4 & 95   & 3.44  \\ 
\midrule
\multirow{2}{*}{Ours} 
& NAR-L            & 369M     & 96.2     & 34    & \textbf{1.09}     \\
& NAR-LP$\dagger$           & 694M     & 71.1     & 34    & 1.30     \\
\bottomrule

\end{tabular}}
\end{table}

\subsubsection{Text to Image Generation}

\begin{table*}[h] \small
\caption{Quantitative evaluation on the GenEval benchmark. $\dagger$ denotes the results are reported by \cite{han2024infinity}.}
\label{tab:geneval}
\centering
\scalebox{1.0}{
\begin{tabular}{ccccccccc}
\hline
Model    & Params & Training Data  & Counting & Two Object & Single Object & Colors & Overall & \begin{tabular}[c]{@{}c@{}}Throughput\\ (img/s)\end{tabular}\\ \hline

LlamaGen-XL$\dagger$~\cite{llamagen}  & 0.8B   & 60M         & -        & 0.34       & -             & -      & 0.32   & 3.40  \\
Chameleon$\dagger$~\cite{team2024chameleon} & 7B   & 1.4B       & -        & -          & -             & -      & 0.39  & 0.09   \\
SDv1.5~\cite{Rombach_2022_LDM}       & 0.9B   & 2B            & \textbf{0.35}     & \textbf{0.38}       & 0.97          & 0.76   & 0.43 & 0.44 \\
NAR-XL       & 0.8B   & \textbf{6M}            & 0.34     & 0.37       & \textbf{0.98}          & \textbf{0.76}   & \textbf{0.43}   & \textbf{15.0} \\ \hline
\end{tabular}}
\end{table*}

To verify the effectiveness of NAR in text-guided image generation, we train a text-guided NAR-XL model and evaluate its performance on the GenEval~\cite{ghosh2024geneval} benchmark, as detailed in Table~\ref{tab:geneval}. The NAR-XL model, trained on merely 6 million publicly available text-image pairs, outperforms LlamaGen-XL~\cite{llamagen} on GenEval benchmark with a notable margin (0.43 \emph{vs.} 0.32), despite \textbf{utilizing just 10\% of the training data and delivering 4.4$\times$ higher throughput}. Moreover, the overall score of the NAR model surpasses that of Chameleon~\cite{team2024chameleon}, an AR visual generation model with 7B parameters trained on 1.4B text-image pairs. Compared to diffusion-based model SDv1.5~\cite{Rombach_2022_LDM}, NAR \textbf{achieves comparable performance while using only $0.4\%$ of the training data.} 
These results underscore the capability of the NAR paradigm to generate high-quality images with minimal training data.
% Qualitative results are presented in Figure~\ref{fig:visualization_parti}.

Quantitative visualizations demonstrating class-conditional image generation, video generation, and text-guided image generation are provided in the supplementary material.

\begin{figure*}[htbp]  
\centering  
\begin{subfigure}{.325\linewidth}  
  \centering  
  \includegraphics[width=1.0\linewidth]{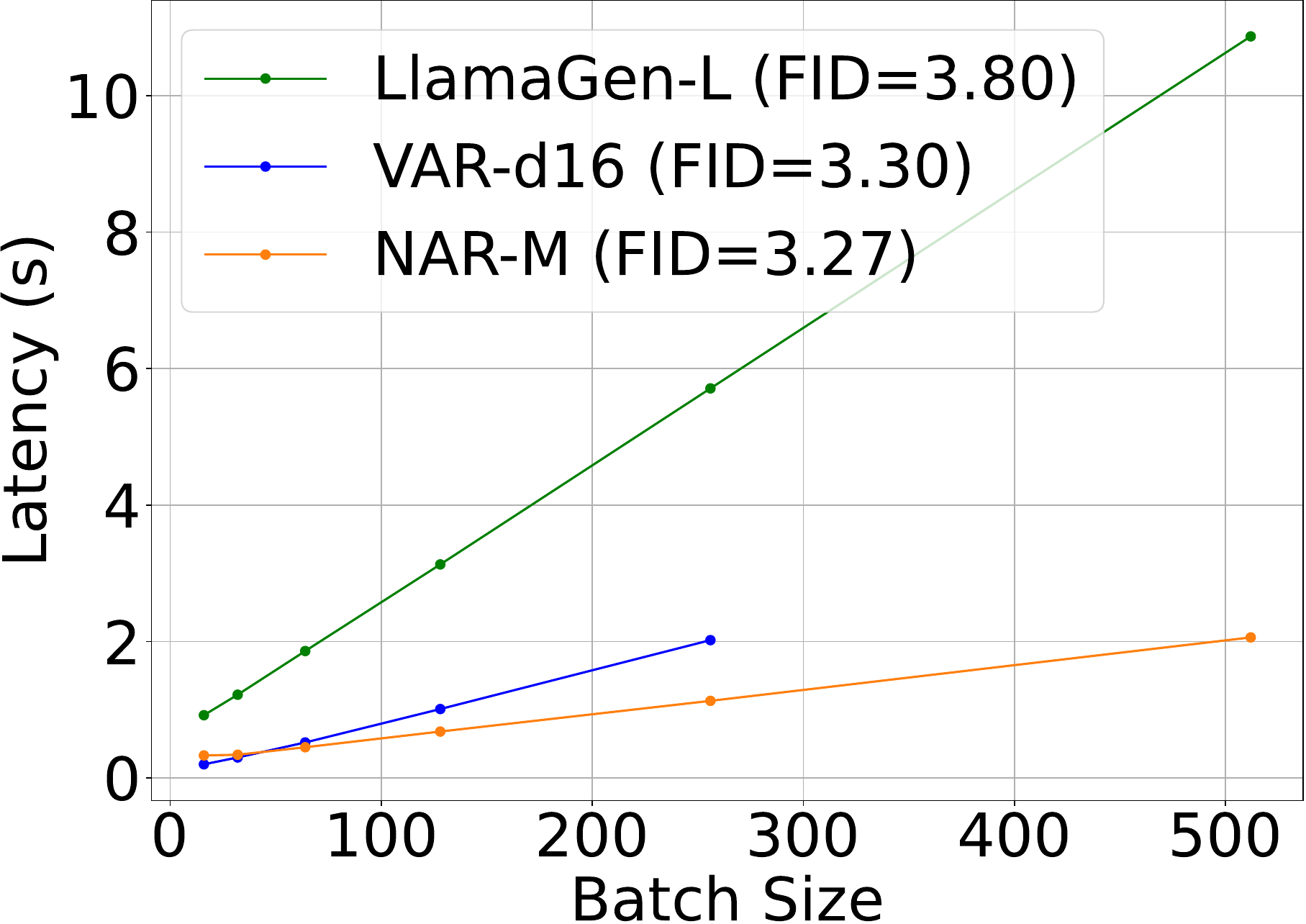}  
  \label{fig:sub1}  
\end{subfigure}%
  \begin{subfigure}{.33\linewidth} 
  \centering  
  \includegraphics[width=1.0\linewidth]{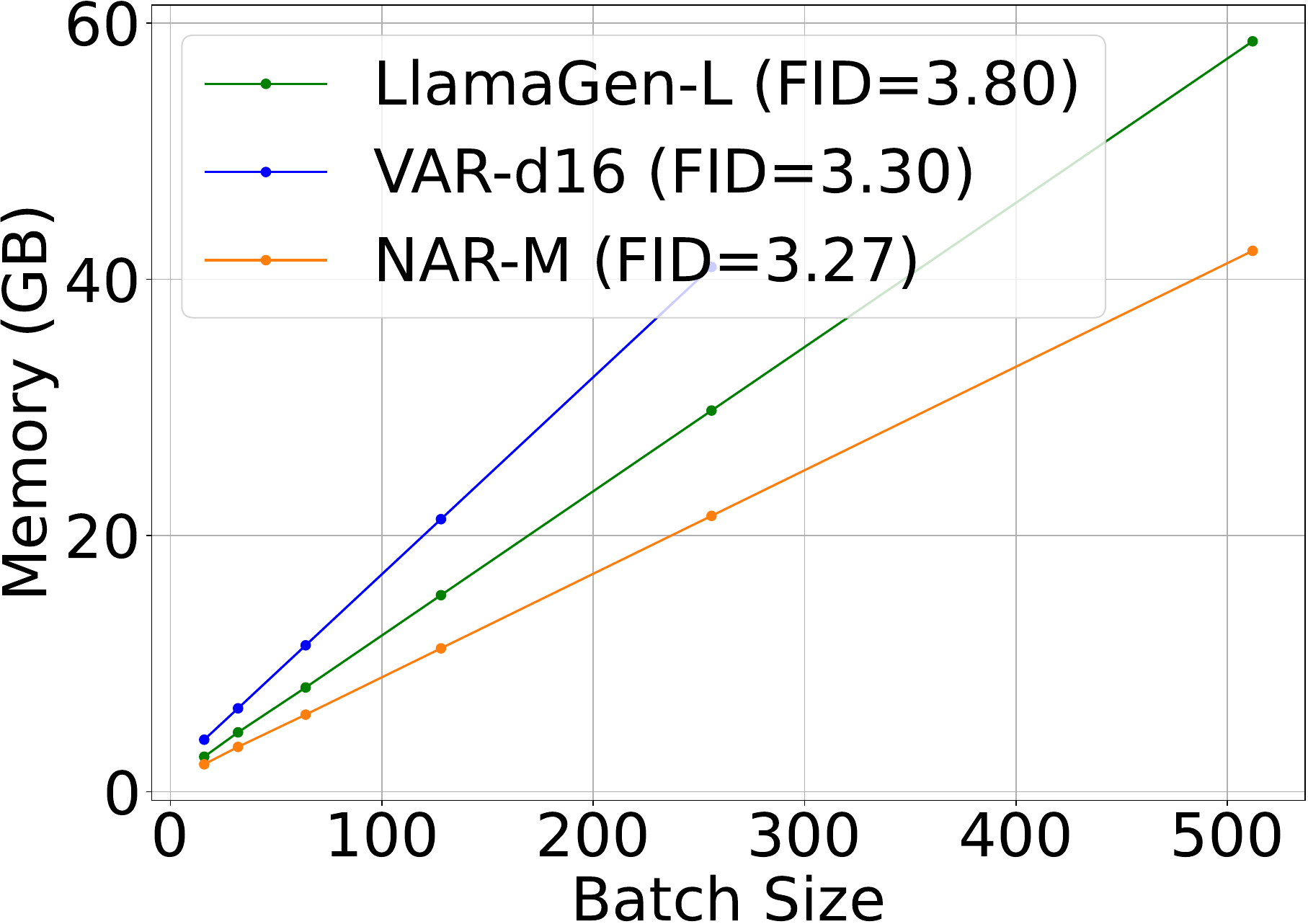}  
  \label{fig:sub2}  
\end{subfigure}%
\begin{subfigure}{.34\linewidth}  
  \centering  
  \includegraphics[width=1.0\linewidth]{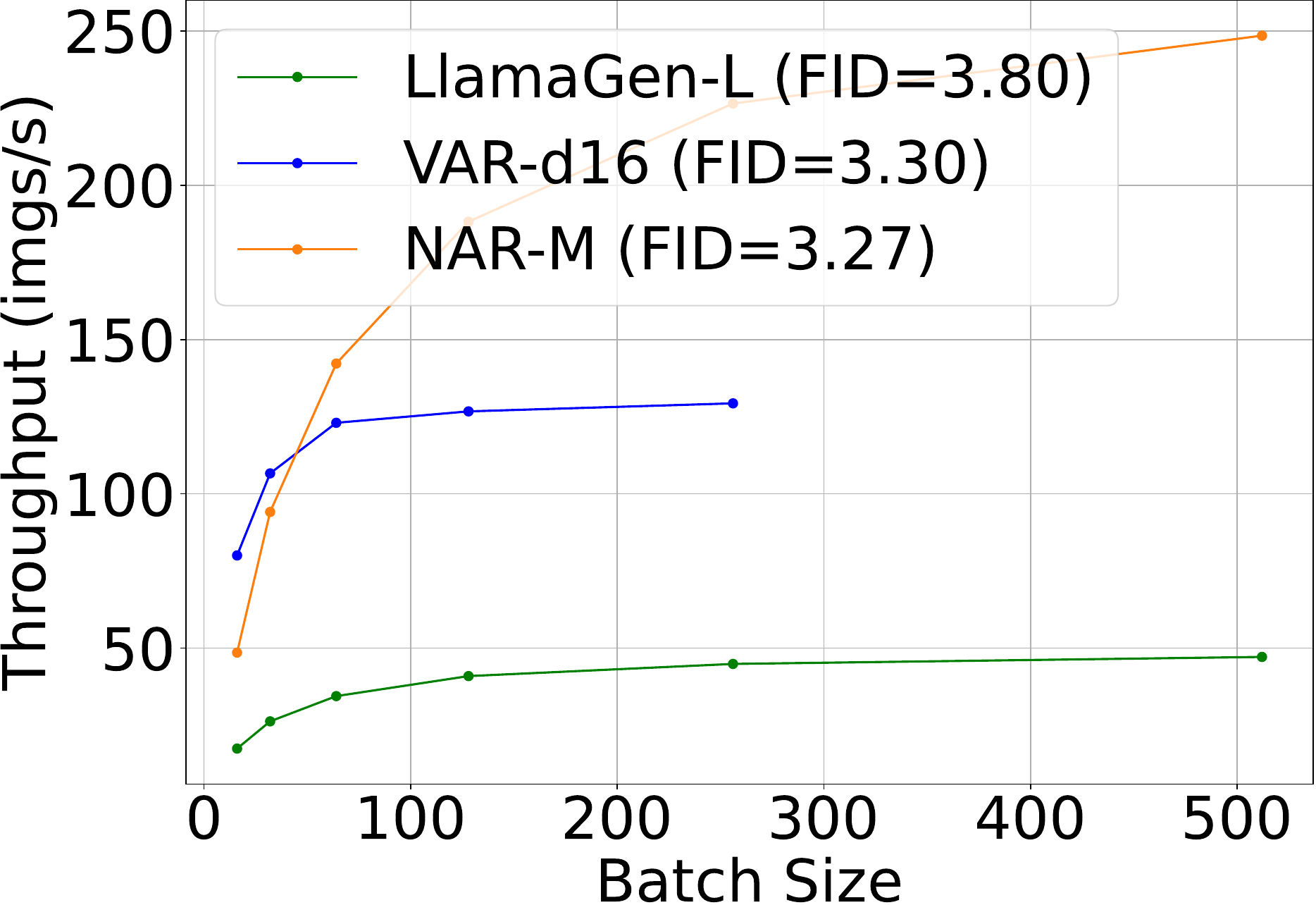}  
  \label{fig:sub3}  
\end{subfigure}%
\caption{Efficiency comparisons between vanilla AR, VAR and the proposed NAR paradigm for visual generation. With a batch size larger than 64, NAR achieves a lower FID with lower latency, lower memory usage and significantly higher throughput.}  
\label{fig:efficiency}  
\end{figure*}

\subsection{Deployment Efficiency} \label{sec:deployment}
In this subsection, we provide a detailed comparison of the efficiency of various autoregressive generation paradigms, evaluating the latency, memory usage, and throughput of models with similar FID scores. As illustrated in Figure~\ref{fig:efficiency}(a), in terms of generation latency, VAR-d16 exhibits lower latency than both NAR-M and LlamaGen-L when the batch size is smaller than 32. This can be attributed to the memory bottleneck during the decoding process in the vanilla next-token AR and NAR paradigms. However, as the batch size increases, NAR-M achieves significantly lower latency. For instance, with a batch size of 256, NAR-M \textbf{reduces latency by 44\%} compared to VAR-d16 (1.13s \emph{vs.} 2.02s). Furthermore, as shown in Figure~\ref{fig:efficiency}(b), NAR-M consistently \textbf{requires less GPU memory} than VAR-d16 for the same batch size, which is due to the shorter sequence length of NAR. Consequently, on the same hardware, NAR paradigm can accommodate a larger batch size, leading to greater throughput, as illustrated in Figure~\ref{fig:efficiency}(c). For example, on an A100 GPU with 80GB of VRAM, VAR-d16 supports a maximum batch size of 256 during inference, yielding a throughput of 129.3 images per second. In contrast, NAR-M supports a batch size of 512, achieving a throughput of 248.5 images per second, which is 92.1\% higher than that of VAR-d16. Compared to the vanilla next-token AR model LlamaGen-L, NAR-M \textbf{provides 5.2$\times$ higher throughput while achieving a superior FID.} 
These results highlight the efficiency advantages of the NAR paradigm in terms of latency, memory usage, and throughput, making it a compelling choice for high-performance and efficient image generation.

\subsection{Ablation Study}

\noindent \textbf{Effect of dimension-oriented decoding heads.}
We assess the effectiveness of the proposed dimension-oriented decoding heads by comparing their performance with a baseline method that employs a single head to predict all adjacent tokens in parallel. For a fair comparison, all models are trained with the same pipeline and hyper-parameters. As demonstrated in Table~\ref{tab:ablate_heads}, NAR-L without dimension-oriented decoding heads significantly underperforms compared to LlamaGen-L with a next-token prediction objective, leading to a substantially higher FID of 66.31. This performance degradation can be attributed to the single decoding head's inability to adequately capture the distinct token distributions at different positions, which may vary considerably for spatially distant tokens. In contrast, NAR-L equipped with the proposed dimension-oriented decoding heads achieves a markedly lower FID of 3.06 while reducing the number of model forward steps by 87.8\% (from 256 steps to 31 steps). This improvement highlights the efficacy of the dimension-oriented heads in enhancing generation quality and efficiency.

\begin{table}[h] \small
\caption{The effect of dimension-oriented decoding heads evaluated on ImageNet $256\times256$ benchmark. Here, ``NTP'' denotes the next-token prediction paradigm. }
\label{tab:ablate_heads}
\centering
\scalebox{1.0}{
\begin{tabular}{cccc}
\hline
Method & Dimension-oriented Heads & Steps & FID$\downarrow$ \\ \hline
LlamaGen-L    & $\times$                 & 256   & 3.80            \\
NAR-L    & $\times$                 & 31    & 66.31               \\
NAR-L    & $\checkmark$             & 31    & \textbf{3.06}   \\ \hline
\end{tabular} }
\end{table}

\noindent \textbf{Effect of mixed logits sampling.}
As illustrated in Figure~\ref{fig:architecture}, overlaps exist between neighboring tokens predicted by different dimension-oriented decoding heads. During inference, these overlapping tokens can be sampled either from the predictions of a single decoding head or from a combination of predictions across multiple decoding heads. We assess the impact of various decoding head configurations on the class-conditional ImageNet $256\times256$ benchmark, as presented in Table~\ref{tab:mixlogits}. The results demonstrate that combining predictions from multiple decoding heads significantly enhances generation quality compared to relying on a single decoding head. This improvement is evidenced by a substantially lower FID of $3.06$ and a higher IS of $263.9$.

\begin{table}[h] \small
\caption{Performance comparison of various decoding head configurations on ImageNet $256\times256$. Here, ``Head'' denotes the decoding heads to use for the overlapped region. Results are obtained with a classifier-free guidance of 2.0. }
\label{tab:mixlogits}
\centering
\begin{tabular}{cccc}
\hline
Model                  & Head   & FID$\downarrow$ & IS$\uparrow$ \\ \hline
\multirow{3}{*}{NAR-L} & Horizontal & 75.91   & 22.36\\
                       & Vertical & 25.73   & 96.08  \\
                       & Mixed   & \textbf{3.06}   & \textbf{263.9}  \\ \hline
\end{tabular}
\end{table}

%% file: sec/5_conclusion.tex
%------------------------------------------------------------------------
\section{Conclusion}
In this paper, we have proposed Neighboring Autoregressive Modeling (NAR), a new "next-neighbor prediction" paradigm for efficient and high-quality visual generation. 
To facilitate the parallel decoding of multiple equidistant tokens, we propose a set of dimension-oriented decoding heads, each responsible for predicting the next token along a mutually orthogonal dimension in the spatial-temporal space. During inference, our approach enables the prediction of all adjacent tokens of a generated token in a single forward step, significantly reducing the number of model forward steps for autoregressive visual generation. Extensive experimental results demonstrate that NAR achieves state-of-the-art generation quality and efficiency trade-off for both image and video generation tasks.

\noindent\textbf{Limitations and future work.}
In this work, we employ a moderate-size image tokenizer to ensure a fair comparison with existing methods~\cite{llamagen, wang2024parallelized}. In the future, we anticipate integrating NAR with more advanced visual tokenizers to further enhance its performance. Moreover, while NAR demonstrates fast and high-quality video generation capabilities, our experiments are currently limited to the class-conditional benchmark UCF-101. We anticipate that training NAR on large-scale video datasets will yield even better video generation results.

%% file: sec/appendix.tex
\begin{center}
	{
		\Large{\textbf{Appendix}}
	}
\end{center}
\appendix
\setcounter{section}{0}
\setcounter{equation}{0}
\setcounter{figure}{0}
\setcounter{table}{0}

\renewcommand\thesection{\Alph{section}}
\renewcommand\thefigure{\Alph{figure}}
\renewcommand\thetable{\Alph{table}}
\renewcommand{\theequation}{\Alph{equation}}
%------------------------------------------------------------------------
\section{More Visualizations}

\begin{figure*}[h]
\centering
\includegraphics[width=0.9\textwidth]{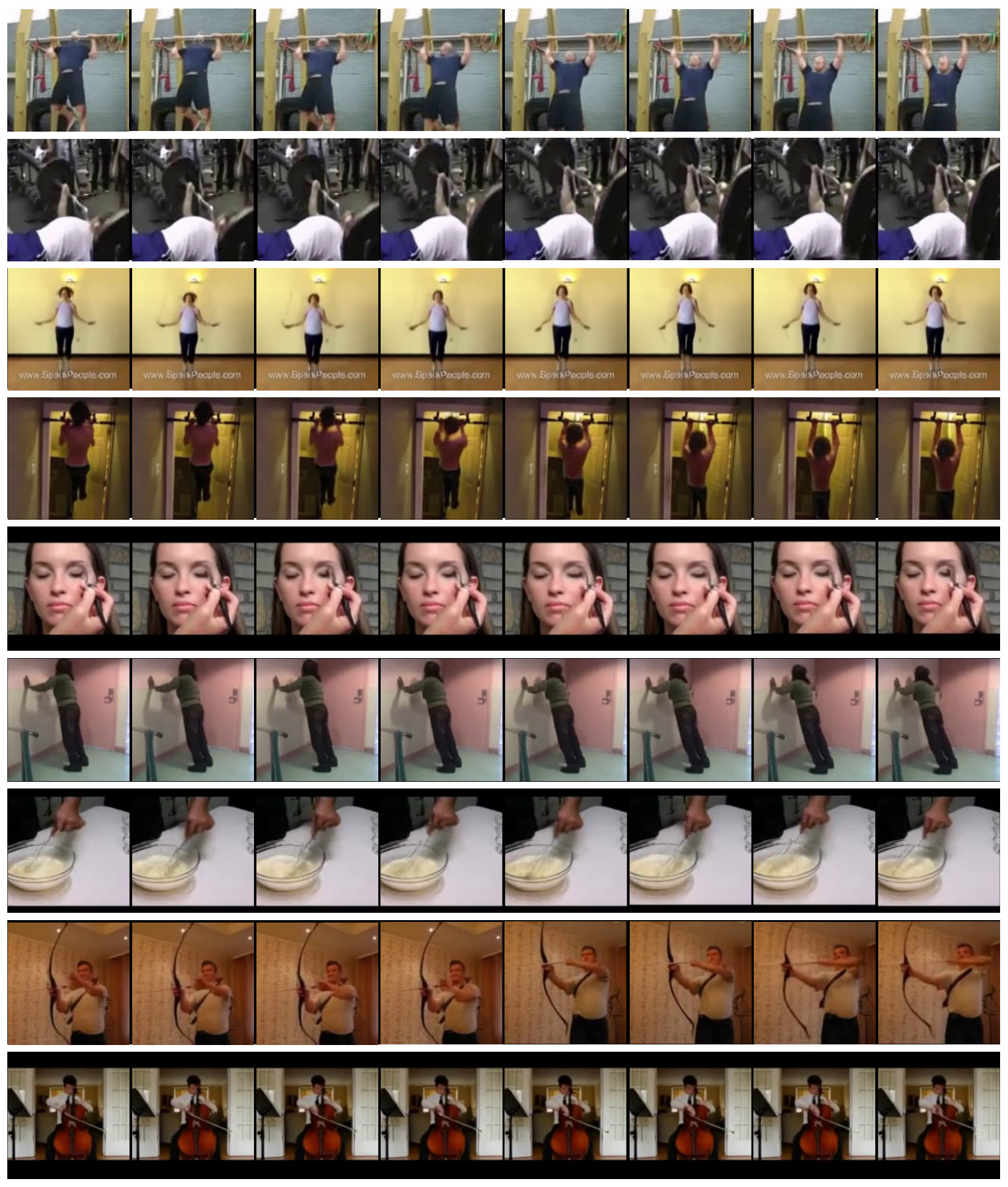}
\caption{ 
\textbf{Video generation samples} on UCF-101 dataset. Each row shows sampled frames from a 16-frame, $128\times128$ resolution sequence generated by NAR-XL across various action categories.}
\label{fig:visualization_ucf}
\end{figure*}

\begin{figure*}[h]
\centering
\includegraphics[width=0.9\textwidth]{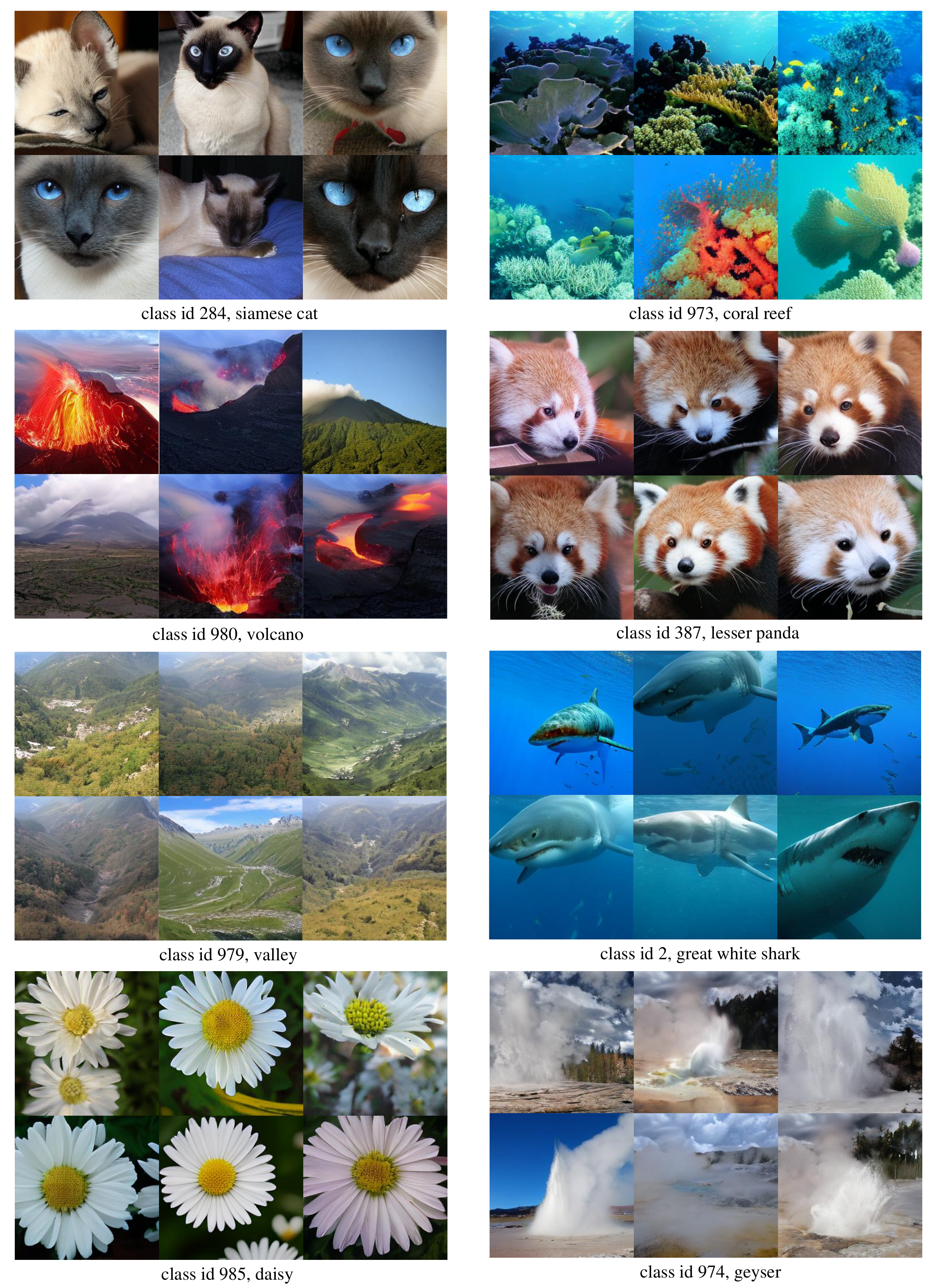}
\caption{ 
\textbf{Class-conditional image generation samples} produced by NAR-XXL on ImageNet $256\times256$.}
\label{fig:visualization_imgnet_1}
\end{figure*}

\begin{figure*}[h]
\centering
\includegraphics[width=0.9\textwidth]{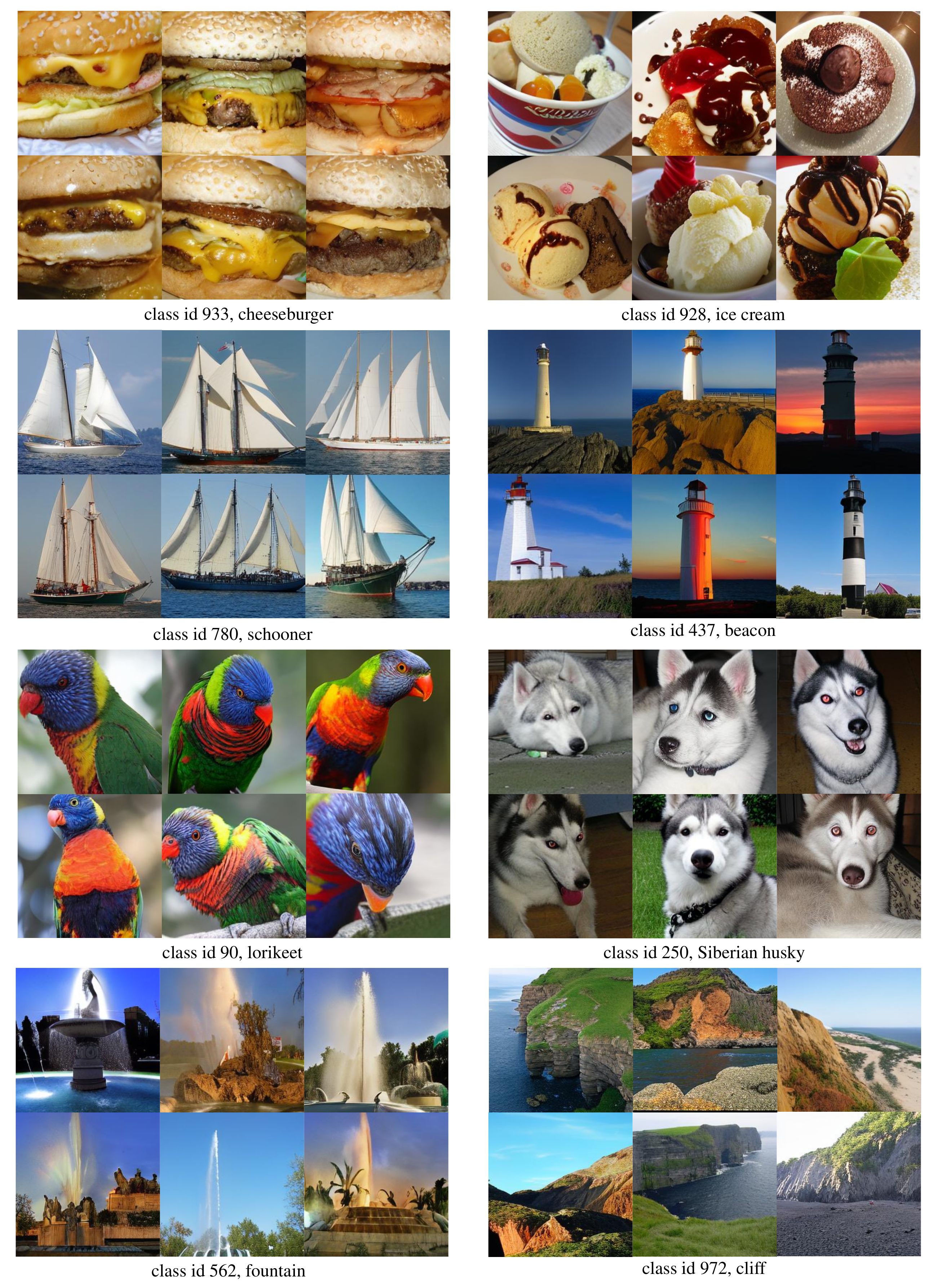}
\caption{ 
\textbf{Class-conditional image generation samples} produced by NAR-XXL on ImageNet $256\times256$.}
\label{fig:visualization_imgnet_2}
\end{figure*}

\begin{figure*}[h]
\centering
\includegraphics[width=0.9\textwidth]{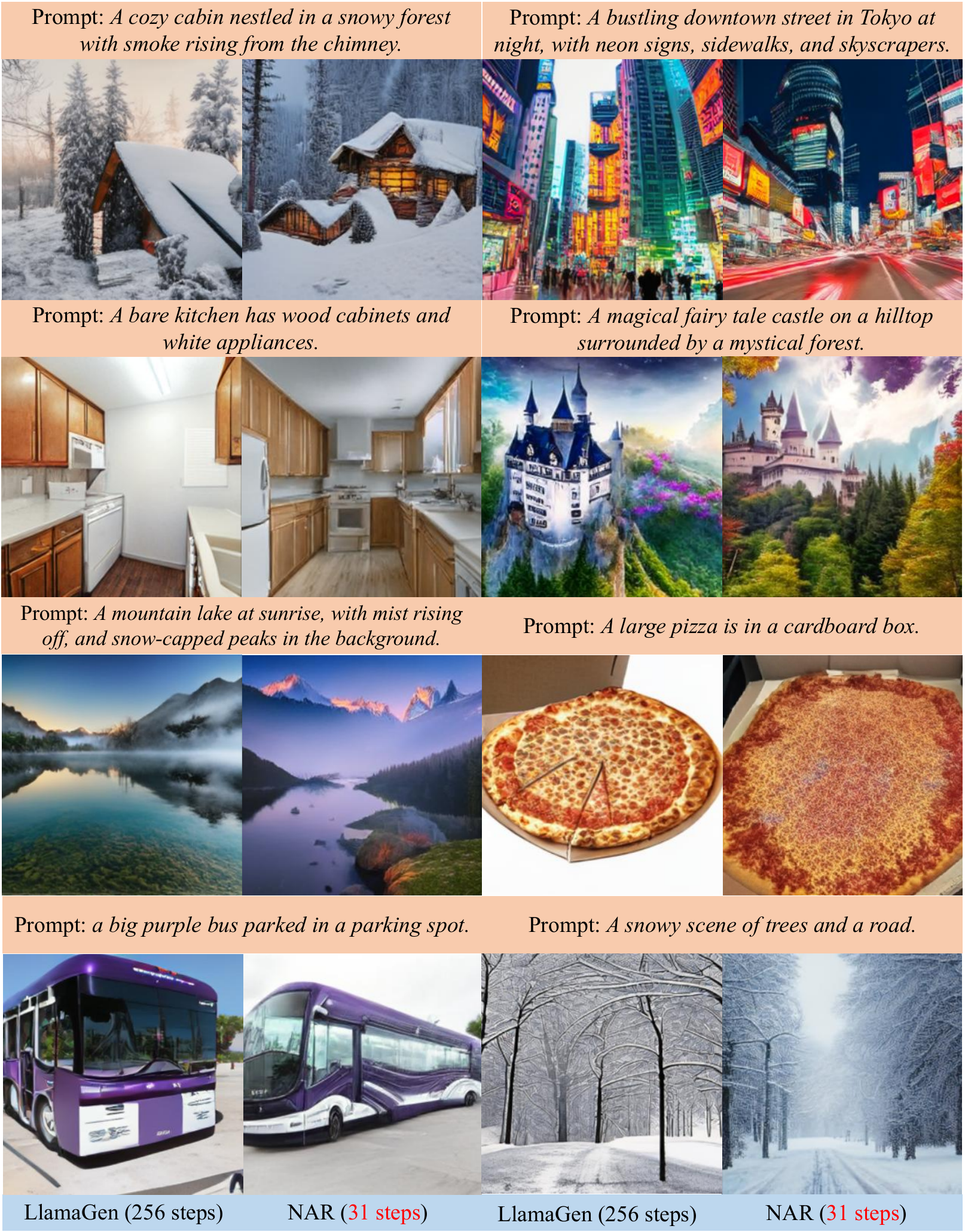}
\caption{ 
\textbf{$256\times256$ text-guided image generation samples} produced by  LlamaGen-XL-Stage1 with next-token prediction paradigm and NAR-XL-Stage1 with next-neighbor prediction paradigm.}
\label{fig:visualization_stage1}
\end{figure*}

\begin{figure*}[h]
\centering
\includegraphics[width=0.9\textwidth]{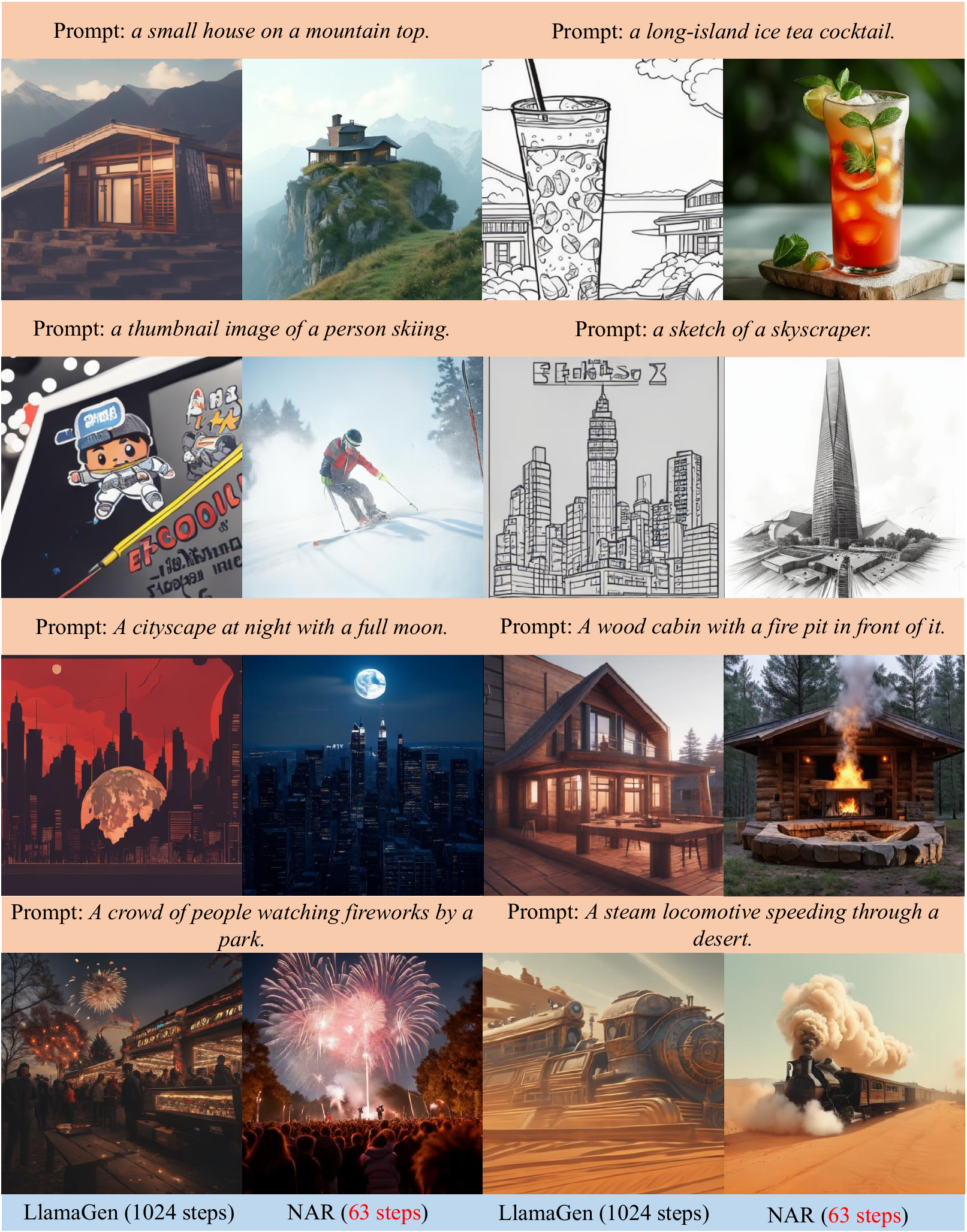}
\caption{ 
\textbf{$512\times512$ text-guided image generation samples} produced by  LlamaGen-XL-Stage2 with next-token prediction paradigm and NAR-XL-Stage2 with next-neighbor prediction paradigm. The text prompts are sampled from Parti prompts.}
\label{fig:visualization_parti}
\end{figure*}

%% file: main.bbl
\begin{thebibliography}{62}
\providecommand{\natexlab}[1]{#1}
\providecommand{\url}[1]{\texttt{#1}}
\expandafter\ifx\csname urlstyle\endcsname\relax
  \providecommand{\doi}[1]{doi: #1}\else
  \providecommand{\doi}{doi: \begingroup \urlstyle{rm}\Url}\fi

\bibitem[lai()]{laioncoco2023}
Laion-coco: 600m synthetic captions from laion-2b-en.
\newblock \emph{https://laion.ai/blog/laion-coco/}.

\bibitem[Bello et~al.(2020)Bello, Zoph, Vaswani, Shlens, and Le]{bello2020attnaug}
Irwan Bello, Barret Zoph, Ashish Vaswani, Jonathon Shlens, and Quoc~V. Le.
\newblock Attention augmented convolutional networks, 2020.

\bibitem[Boutell(1997)]{boutell1997png}
Thomas Boutell.
\newblock Png (portable network graphics) specification version 1.0.
\newblock Technical report, 1997.

\bibitem[Canny(1986)]{canny1986computational}
John Canny.
\newblock A computational approach to edge detection.
\newblock \emph{IEEE Transactions on pattern analysis and machine intelligence}, \penalty0 (6):\penalty0 679--698, 1986.

\bibitem[Carion et~al.(2020)Carion, Massa, Synnaeve, Usunier, Kirillov, and Zagoruyko]{carion2020endtoend}
Nicolas Carion, Francisco Massa, Gabriel Synnaeve, Nicolas Usunier, Alexander Kirillov, and Sergey Zagoruyko.
\newblock End-to-end object detection with transformers, 2020.

\bibitem[Chen et~al.(2017)Chen, Mishra, Rohaninejad, and Abbeel]{xi2017pixelsnail}
Xi Chen, Nikhil Mishra, Mostafa Rohaninejad, and Pieter Abbeel.
\newblock Pixelsnail: An improved autoregressive generative model, 2017.

\bibitem[Chen et~al.(2025)Chen, Wu, Liu, Pan, Liu, Xie, Yu, and Ruan]{janus-pro}
Xiaokang Chen, Zhiyu Wu, Xingchao Liu, Zizheng Pan, Wen Liu, Zhenda Xie, Xingkai Yu, and Chong Ruan.
\newblock Janus-pro: Unified multimodal understanding and generation with data and model scaling.
\newblock \emph{arXiv preprint arXiv:2501.17811}, 2025.

\bibitem[Chung et~al.(2024)Chung, Hou, Longpre, Zoph, Tay, Fedus, Li, Wang, Dehghani, Brahma, et~al.]{chung2024flant5}
Hyung~Won Chung, Le Hou, Shayne Longpre, Barret Zoph, Yi Tay, William Fedus, Yunxuan Li, Xuezhi Wang, Mostafa Dehghani, Siddhartha Brahma, et~al.
\newblock Scaling instruction-finetuned language models.
\newblock \emph{Journal of Machine Learning Research}, 25\penalty0 (70):\penalty0 1--53, 2024.

\bibitem[Dabov et~al.(2007)Dabov, Foi, Katkovnik, and Egiazarian]{dabov2007imagedenoising}
Kostadin Dabov, Alessandro Foi, Vladimir Katkovnik, and Karen Egiazarian.
\newblock Image denoising by sparse 3-d transform-domain collaborative filtering.
\newblock \emph{IEEE Transactions on image processing}, 16\penalty0 (8):\penalty0 2080--2095, 2007.

\bibitem[Deng et~al.(2024)Deng, Pan, Diao, Luo, Cui, Lu, Shan, Qi, and Wang]{deng2024nova}
Haoge Deng, Ting Pan, Haiwen Diao, Zhengxiong Luo, Yufeng Cui, Huchuan Lu, Shiguang Shan, Yonggang Qi, and Xinlong Wang.
\newblock Autoregressive video generation without vector quantization.
\newblock \emph{arXiv preprint arXiv:2412.14169}, 2024.

\bibitem[Dhariwal and Nichol(2021)]{dhariwal2021adm}
Prafulla Dhariwal and Alexander Nichol.
\newblock Diffusion models beat gans on image synthesis.
\newblock \emph{Advances in neural information processing systems}, 34:\penalty0 8780--8794, 2021.

\bibitem[Dosovitskiy(2020)]{dosovitskiy2020vit}
Alexey Dosovitskiy.
\newblock An image is worth 16x16 words: Transformers for image recognition at scale.
\newblock \emph{arXiv preprint arXiv:2010.11929}, 2020.

\bibitem[Esser et~al.(2021)Esser, Rombach, and Ommer]{esser2021taming}
Patrick Esser, Robin Rombach, and Bjorn Ommer.
\newblock Taming transformers for high-resolution image synthesis.
\newblock In \emph{Proceedings of the IEEE/CVF conference on computer vision and pattern recognition}, pages 12873--12883, 2021.

\bibitem[Ge et~al.(2022)Ge, Hayes, Yang, Yin, Pang, Jacobs, Huang, and Parikh]{ge2022long}
Songwei Ge, Thomas Hayes, Harry Yang, Xi Yin, Guan Pang, David Jacobs, Jia-Bin Huang, and Devi Parikh.
\newblock Long video generation with time-agnostic vqgan and time-sensitive transformer.
\newblock In \emph{European Conference on Computer Vision}, pages 102--118. Springer, 2022.

\bibitem[Ghosh et~al.(2024)Ghosh, Hajishirzi, and Schmidt]{ghosh2024geneval}
Dhruba Ghosh, Hannaneh Hajishirzi, and Ludwig Schmidt.
\newblock Geneval: An object-focused framework for evaluating text-to-image alignment.
\newblock \emph{Advances in Neural Information Processing Systems}, 36, 2024.

\bibitem[Han et~al.(2024)Han, Liu, Jiang, Yan, Zhang, Yuan, Peng, and Liu]{han2024infinity}
Jian Han, Jinlai Liu, Yi Jiang, Bin Yan, Yuqi Zhang, Zehuan Yuan, Bingyue Peng, and Xiaobing Liu.
\newblock Infinity: Scaling bitwise autoregressive modeling for high-resolution image synthesis.
\newblock \emph{arXiv preprint arXiv:2412.04431}, 2024.

\bibitem[He et~al.(2015)He, Zhang, Ren, and Sun]{he2015deep}
Kaiming He, Xiangyu Zhang, Shaoqing Ren, and Jian Sun.
\newblock Deep residual learning for image recognition. arxiv e-prints.
\newblock \emph{arXiv preprint arXiv:1512.03385}, 10, 2015.

\bibitem[He et~al.(2017)He, Gkioxari, Doll{\'a}r, and Girshick]{he2017mask}
Kaiming He, Georgia Gkioxari, Piotr Doll{\'a}r, and Ross Girshick.
\newblock Mask r-cnn.
\newblock In \emph{Proceedings of the IEEE international conference on computer vision}, pages 2961--2969, 2017.

\bibitem[He et~al.(2024)He, Chen, He, He, Zhou, Zhang, and Zhuang]{he2024zipar}
Yefei He, Feng Chen, Yuanyu He, Shaoxuan He, Hong Zhou, Kaipeng Zhang, and Bohan Zhuang.
\newblock Zipar: Accelerating autoregressive image generation through spatial locality.
\newblock \emph{arXiv preprint arXiv:2412.04062}, 2024.

\bibitem[Hong et~al.(2022)Hong, Ding, Zheng, Liu, and Tang]{hong2022cogvideo}
Wenyi Hong, Ming Ding, Wendi Zheng, Xinghan Liu, and Jie Tang.
\newblock Cogvideo: Large-scale pretraining for text-to-video generation via transformers.
\newblock \emph{arXiv preprint arXiv:2205.15868}, 2022.

\bibitem[Krizhevsky et~al.(2012)Krizhevsky, Sutskever, and Hinton]{krizhevsky2012imagenet}
Alex Krizhevsky, Ilya Sutskever, and Geoffrey~E Hinton.
\newblock Imagenet classification with deep convolutional neural networks.
\newblock \emph{NeurIPS}, 2012.

\bibitem[Kuznetsova et~al.(2020)Kuznetsova, Rom, Alldrin, Uijlings, Krasin, Pont-Tuset, Kamali, Popov, Malloci, Kolesnikov, et~al.]{kuznetsova2020openimages}
Alina Kuznetsova, Hassan Rom, Neil Alldrin, Jasper Uijlings, Ivan Krasin, Jordi Pont-Tuset, Shahab Kamali, Stefan Popov, Matteo Malloci, Alexander Kolesnikov, et~al.
\newblock The open images dataset v4: Unified image classification, object detection, and visual relationship detection at scale.
\newblock \emph{International journal of computer vision}, 128\penalty0 (7):\penalty0 1956--1981, 2020.

\bibitem[LeCun et~al.(1989)LeCun, Boser, Denker, Henderson, Howard, Hubbard, and Jackel]{lenet}
Y. LeCun, B. Boser, J.~S. Denker, D. Henderson, R.~E. Howard, W. Hubbard, and L.~D. Jackel.
\newblock Backpropagation applied to handwritten zip code recognition.
\newblock \emph{Neural Computation}, 1\penalty0 (4):\penalty0 541--551, 1989.

\bibitem[Li et~al.(2024{\natexlab{a}})Li, Tian, Li, Deng, and He]{li2024mar}
Tianhong Li, Yonglong Tian, He Li, Mingyang Deng, and Kaiming He.
\newblock Autoregressive image generation without vector quantization.
\newblock \emph{arXiv preprint arXiv:2406.11838}, 2024{\natexlab{a}}.

\bibitem[Li et~al.(2024{\natexlab{b}})Li, Qiu, Chen, Kuen, Lin, Singh, and Raj]{li2024controlvar}
Xiang Li, Kai Qiu, Hao Chen, Jason Kuen, Zhe Lin, Rita Singh, and Bhiksha Raj.
\newblock Controlvar: Exploring controllable visual autoregressive modeling.
\newblock \emph{arXiv preprint arXiv:2406.09750}, 2024{\natexlab{b}}.

\bibitem[Lin et~al.(2021)Lin, Pagnucco, and Song]{lin2021edge}
Han Lin, Maurice Pagnucco, and Yang Song.
\newblock Edge guided progressively generative image outpainting.
\newblock In \emph{Proceedings of the IEEE/CVF conference on computer vision and pattern recognition}, pages 806--815, 2021.

\bibitem[Liu et~al.(2024)Liu, Zhao, Zhuo, Lin, Qiao, Li, and Gao]{liu2024lumina}
Dongyang Liu, Shitian Zhao, Le Zhuo, Weifeng Lin, Yu Qiao, Hongsheng Li, and Peng Gao.
\newblock Lumina-mgpt: Illuminate flexible photorealistic text-to-image generation with multimodal generative pretraining.
\newblock \emph{arXiv preprint arXiv:2408.02657}, 2024.

\bibitem[Liu et~al.(2021)Liu, Lin, Cao, Hu, Wei, Zhang, Lin, and Guo]{swin}
Ze Liu, Yutong Lin, Yue Cao, Han Hu, Yixuan Wei, Zheng Zhang, Stephen Lin, and Baining Guo.
\newblock Swin transformer: Hierarchical vision transformer using shifted windows, 2021.

\bibitem[Long et~al.(2014)Long, Shelhamer, Darrell, and Berkeley]{long2014fully}
Jonathan Long, Evan Shelhamer, Trevor Darrell, and UC Berkeley.
\newblock Fully convolutional networks for semantic segmentation. arxiv 2015.
\newblock \emph{arXiv preprint arXiv:1411.4038}, 2014.

\bibitem[Lu et~al.(2024)Lu, Pang, Xiao, Zhu, Xia, and Zhang]{lu2024ensemble}
Jinliang Lu, Ziliang Pang, Min Xiao, Yaochen Zhu, Rui Xia, and Jiajun Zhang.
\newblock Merge, ensemble, and cooperate! a survey on collaborative strategies in the era of large language models.
\newblock \emph{arXiv preprint arXiv:2407.06089}, 2024.

\bibitem[Luo et~al.(2023)Luo, Chen, Zhang, Huang, Wang, Shen, Zhao, Zhou, and Tan]{luo2023videofusion}
Zhengxiong Luo, Dayou Chen, Yingya Zhang, Yan Huang, Liang Wang, Yujun Shen, Deli Zhao, Jingren Zhou, and Tieniu Tan.
\newblock Videofusion: Decomposed diffusion models for high-quality video generation.
\newblock \emph{arXiv preprint arXiv:2303.08320}, 2023.

\bibitem[Peebles and Xie(2023)]{peebles2023scalable}
William Peebles and Saining Xie.
\newblock Scalable diffusion models with transformers.
\newblock In \emph{Proceedings of the IEEE/CVF International Conference on Computer Vision}, pages 4195--4205, 2023.

\bibitem[Podell et~al.(2023)Podell, English, Lacey, Blattmann, Dockhorn, M{\"u}ller, Penna, and Rombach]{sdxl}
Dustin Podell, Zion English, Kyle Lacey, Andreas Blattmann, Tim Dockhorn, Jonas M{\"u}ller, Joe Penna, and Robin Rombach.
\newblock Sdxl: Improving latent diffusion models for high-resolution image synthesis.
\newblock \emph{arXiv preprint arXiv:2307.01952}, 2023.

\bibitem[Radford(2018)]{radford2018improving}
Alec Radford.
\newblock Improving language understanding by generative pre-training.
\newblock 2018.

\bibitem[Radford et~al.(2019)Radford, Wu, Child, Luan, Amodei, Sutskever, et~al.]{radford2019language}
Alec Radford, Jeffrey Wu, Rewon Child, David Luan, Dario Amodei, Ilya Sutskever, et~al.
\newblock Language models are unsupervised multitask learners.
\newblock \emph{OpenAI blog}, 1\penalty0 (8):\penalty0 9, 2019.

\bibitem[Redmon et~al.(2015)Redmon, Divvala, Girshick, and Farhadi]{redmon2015you}
Joseph Redmon, Santosh Divvala, Ross Girshick, and Ali Farhadi.
\newblock You only look once: unified, real-time object detection (2015).
\newblock \emph{arXiv preprint arXiv:1506.02640}, 825, 2015.

\bibitem[Rombach et~al.(2022)Rombach, Blattmann, Lorenz, Esser, and Ommer]{Rombach_2022_LDM}
Robin Rombach, Andreas Blattmann, Dominik Lorenz, Patrick Esser, and Bj\"orn Ommer.
\newblock High-resolution image synthesis with latent diffusion models.
\newblock In \emph{Proceedings of the IEEE/CVF Conference on Computer Vision and Pattern Recognition (CVPR)}, pages 10684--10695, 2022.

\bibitem[Sabini and Rusak(2018)]{sabini2018painting}
Mark Sabini and Gili Rusak.
\newblock Painting outside the box: Image outpainting with gans.
\newblock \emph{arXiv preprint arXiv:1808.08483}, 2018.

\bibitem[Simonyan(2014)]{simonyan2014very}
Karen Simonyan.
\newblock Very deep convolutional networks for large-scale image recognition.
\newblock \emph{arXiv preprint arXiv:1409.1556}, 2014.

\bibitem[Singer et~al.(2022)Singer, Polyak, Hayes, Yin, An, Zhang, Hu, Yang, Ashual, Gafni, et~al.]{singer2022make}
Uriel Singer, Adam Polyak, Thomas Hayes, Xi Yin, Jie An, Songyang Zhang, Qiyuan Hu, Harry Yang, Oron Ashual, Oran Gafni, et~al.
\newblock Make-a-video: Text-to-video generation without text-video data.
\newblock \emph{arXiv preprint arXiv:2209.14792}, 2022.

\bibitem[Skorokhodov et~al.(2024)Skorokhodov, Menapace, Siarohin, and Tulyakov]{skorokhodov2024hierarchical}
Ivan Skorokhodov, Willi Menapace, Aliaksandr Siarohin, and Sergey Tulyakov.
\newblock Hierarchical patch diffusion models for high-resolution video generation.
\newblock In \emph{Proceedings of the IEEE/CVF Conference on Computer Vision and Pattern Recognition}, pages 7569--7579, 2024.

\bibitem[Sobel and Feldman(1973)]{Sobel1973}
I. Sobel and G. Feldman.
\newblock \emph{A 3x3 Isotropic Gradient Operator for Image Processing}, pages 271--272.
\newblock 1973.

\bibitem[Soomro(2012)]{soomro2012ucf101}
K Soomro.
\newblock Ucf101: A dataset of 101 human actions classes from videos in the wild.
\newblock \emph{arXiv preprint arXiv:1212.0402}, 2012.

\bibitem[Sun et~al.(2024)Sun, Jiang, Chen, Zhang, Peng, Luo, and Yuan]{llamagen}
Peize Sun, Yi Jiang, Shoufa Chen, Shilong Zhang, Bingyue Peng, Ping Luo, and Zehuan Yuan.
\newblock Autoregressive model beats diffusion: Llama for scalable image generation.
\newblock \emph{arXiv preprint arXiv:2406.06525}, 2024.

\bibitem[Team(2024)]{team2024chameleon}
Chameleon Team.
\newblock Chameleon: Mixed-modal early-fusion foundation models.
\newblock \emph{arXiv preprint arXiv:2405.09818}, 2024.

\bibitem[Teng et~al.(2024)Teng, Shi, Liu, Ning, Dai, Wang, Li, and Liu]{jacobi}
Yao Teng, Han Shi, Xian Liu, Xuefei Ning, Guohao Dai, Yu Wang, Zhenguo Li, and Xihui Liu.
\newblock Accelerating auto-regressive text-to-image generation with training-free speculative jacobi decoding.
\newblock \emph{arXiv preprint arXiv:2410.01699}, 2024.

\bibitem[Tian et~al.(2024)Tian, Jiang, Yuan, Peng, and Wang]{tian2024var}
Keyu Tian, Yi Jiang, Zehuan Yuan, Bingyue Peng, and Liwei Wang.
\newblock Visual autoregressive modeling: Scalable image generation via next-scale prediction.
\newblock 2024.

\bibitem[Tomasi and Manduchi(1998)]{tomasi1998bilateral}
Carlo Tomasi and Roberto Manduchi.
\newblock Bilateral filtering for gray and color images.
\newblock In \emph{Sixth international conference on computer vision (IEEE Cat. No. 98CH36271)}, pages 839--846. IEEE, 1998.

\bibitem[Touvron et~al.(2023{\natexlab{a}})Touvron, Lavril, Izacard, Martinet, Lachaux, Lacroix, Rozi{\`e}re, Goyal, Hambro, Azhar, et~al.]{llama}
Hugo Touvron, Thibaut Lavril, Gautier Izacard, Xavier Martinet, Marie-Anne Lachaux, Timoth{\'e}e Lacroix, Baptiste Rozi{\`e}re, Naman Goyal, Eric Hambro, Faisal Azhar, et~al.
\newblock Llama: Open and efficient foundation language models.
\newblock \emph{arXiv preprint arXiv:2302.13971}, 2023{\natexlab{a}}.

\bibitem[Touvron et~al.(2023{\natexlab{b}})Touvron, Martin, Stone, Albert, Almahairi, Babaei, Bashlykov, Batra, Bhargava, Bhosale, et~al.]{llama2}
Hugo Touvron, Louis Martin, Kevin Stone, Peter Albert, Amjad Almahairi, Yasmine Babaei, Nikolay Bashlykov, Soumya Batra, Prajjwal Bhargava, Shruti Bhosale, et~al.
\newblock Llama 2: Open foundation and fine-tuned chat models.
\newblock \emph{arXiv preprint arXiv:2307.09288}, 2023{\natexlab{b}}.

\bibitem[Unterthiner et~al.(2018)Unterthiner, Van~Steenkiste, Kurach, Marinier, Michalski, and Gelly]{unterthiner2018towards}
Thomas Unterthiner, Sjoerd Van~Steenkiste, Karol Kurach, Raphael Marinier, Marcin Michalski, and Sylvain Gelly.
\newblock Towards accurate generative models of video: A new metric \& challenges.
\newblock \emph{arXiv preprint arXiv:1812.01717}, 2018.

\bibitem[Van~den Oord et~al.(2016)Van~den Oord, Kalchbrenner, Espeholt, Vinyals, Graves, et~al.]{van2016pixelcnn}
Aaron Van~den Oord, Nal Kalchbrenner, Lasse Espeholt, Oriol Vinyals, Alex Graves, et~al.
\newblock Conditional image generation with pixelcnn decoders.
\newblock \emph{Advances in neural information processing systems}, 29, 2016.

\bibitem[Wallace(1991)]{wallace1991jpeg}
Gregory~K Wallace.
\newblock The jpeg still picture compression standard.
\newblock \emph{Communications of the ACM}, 34\penalty0 (4):\penalty0 30--44, 1991.

\bibitem[Wang et~al.(2024{\natexlab{a}})Wang, Suri, Ren, Chen, and Shrivastava]{wang2024larp}
Hanyu Wang, Saksham Suri, Yixuan Ren, Hao Chen, and Abhinav Shrivastava.
\newblock Larp: Tokenizing videos with a learned autoregressive generative prior.
\newblock \emph{arXiv preprint arXiv:2410.21264}, 2024{\natexlab{a}}.

\bibitem[Wang et~al.(2024{\natexlab{b}})Wang, Jiang, Yuan, Peng, Wu, and Jiang]{wang2024omnitokenizer}
Junke Wang, Yi Jiang, Zehuan Yuan, Binyue Peng, Zuxuan Wu, and Yu-Gang Jiang.
\newblock Omnitokenizer: A joint image-video tokenizer for visual generation.
\newblock \emph{arXiv preprint arXiv:2406.09399}, 2024{\natexlab{b}}.

\bibitem[Wang et~al.(2024{\natexlab{c}})Wang, Bai, Tan, Wang, Fan, Bai, Chen, Liu, Wang, Ge, et~al.]{wang2024qwen2vl}
Peng Wang, Shuai Bai, Sinan Tan, Shijie Wang, Zhihao Fan, Jinze Bai, Keqin Chen, Xuejing Liu, Jialin Wang, Wenbin Ge, et~al.
\newblock Qwen2-vl: Enhancing vision-language model's perception of the world at any resolution.
\newblock \emph{arXiv preprint arXiv:2409.12191}, 2024{\natexlab{c}}.

\bibitem[Wang et~al.(2024{\natexlab{d}})Wang, Zhang, Luo, Sun, Cui, Wang, Zhang, Wang, Li, Yu, et~al.]{emu3}
Xinlong Wang, Xiaosong Zhang, Zhengxiong Luo, Quan Sun, Yufeng Cui, Jinsheng Wang, Fan Zhang, Yueze Wang, Zhen Li, Qiying Yu, et~al.
\newblock Emu3: Next-token prediction is all you need.
\newblock \emph{arXiv preprint arXiv:2409.18869}, 2024{\natexlab{d}}.

\bibitem[Wang et~al.(2024{\natexlab{e}})Wang, Ren, Lin, Han, Guo, Yang, Zou, Feng, and Liu]{wang2024parallelized}
Yuqing Wang, Shuhuai Ren, Zhijie Lin, Yujin Han, Haoyuan Guo, Zhenheng Yang, Difan Zou, Jiashi Feng, and Xihui Liu.
\newblock Parallelized autoregressive visual generation.
\newblock \emph{arXiv preprint arXiv:2412.15119}, 2024{\natexlab{e}}.

\bibitem[Waswani et~al.(2017)Waswani, Shazeer, Parmar, Uszkoreit, Jones, Gomez, Kaiser, and Polosukhin]{waswani2017attention}
A Waswani, N Shazeer, N Parmar, J Uszkoreit, L Jones, A Gomez, L Kaiser, and I Polosukhin.
\newblock Attention is all you need.
\newblock In \emph{NIPS}, 2017.

\bibitem[Yu et~al.(2023{\natexlab{a}})Yu, Cheng, Sohn, Lezama, Zhang, Chang, Hauptmann, Yang, Hao, Essa, et~al.]{yu2023magvit}
Lijun Yu, Yong Cheng, Kihyuk Sohn, Jos{\'e} Lezama, Han Zhang, Huiwen Chang, Alexander~G Hauptmann, Ming-Hsuan Yang, Yuan Hao, Irfan Essa, et~al.
\newblock Magvit: Masked generative video transformer.
\newblock In \emph{Proceedings of the IEEE/CVF Conference on Computer Vision and Pattern Recognition}, pages 10459--10469, 2023{\natexlab{a}}.

\bibitem[Yu et~al.(2023{\natexlab{b}})Yu, Lezama, Gundavarapu, Versari, Sohn, Minnen, Cheng, Birodkar, Gupta, Gu, et~al.]{yu2023language}
Lijun Yu, Jos{\'e} Lezama, Nitesh~B Gundavarapu, Luca Versari, Kihyuk Sohn, David Minnen, Yong Cheng, Vighnesh Birodkar, Agrim Gupta, Xiuye Gu, et~al.
\newblock Language model beats diffusion--tokenizer is key to visual generation.
\newblock \emph{arXiv preprint arXiv:2310.05737}, 2023{\natexlab{b}}.

\bibitem[Zhang et~al.(2022)Zhang, Zhang, Zhao, Chen, Arik, and Pfister]{zhang2022nested}
Zizhao Zhang, Han Zhang, Long Zhao, Ting Chen, Sercan~{\"O} Arik, and Tomas Pfister.
\newblock Nested hierarchical transformer: Towards accurate, data-efficient and interpretable visual understanding.
\newblock In \emph{AAAI}, 2022.

\end{thebibliography}
